\documentclass[sigconf]{acmart}

\usepackage{amsmath}
\usepackage{amsthm}
\usepackage{algorithm}
\usepackage[noend]{algpseudocode}
\usepackage{pgfgantt}
\usepackage{xcolor}
\usepackage{soul}

\usepackage[font=small, skip=0.5ex]{caption}

\usepackage{graphicx}
\graphicspath{ {figures/} }
\usepackage{microtype}
\usepackage{subcaption}
\usepackage{tabularx}

\usepackage{subfiles}
\usepackage{xspace}

\usepackage{enumitem}
\usepackage{setspace}
\usepackage{etaremune}
\usepackage{array}
\usepackage[table]{colortbl}

\usepackage[compact]{titlesec}
\usepackage{pgfplots}
\usepgfplotslibrary{fillbetween}	

\allowdisplaybreaks

\DeclareMathOperator*{\argmax}{argmax}

\AtBeginDocument{%
  \providecommand\BibTeX{{%
    \normalfont B\kern-0.5em{\scshape i\kern-0.25em b}\kern-0.8em\TeX}}}

\setcopyright{none}
\acmConference{}
\acmBooktitle{}
\acmDOI{}
\acmPrice{}
\acmISBN{}
\copyrightyear{}

\renewcommand\footnotetextcopyrightpermission[1]{} 
\newcommand{\sysname}{FaiR-IoT\xspace}

\begin{document}

\title{FaiR-IoT: Fairness-aware Human-in-the-Loop Reinforcement Learning for Harnessing Human Variability in Personalized IoT}

\author{Salma Elmalaki}
\email{salma.elmalaki@uci.edu}
\orcid{1234-5678-9012}
\affiliation{%
  \institution{University of California, Irvine}
}

\renewcommand{\shortauthors}{Salma Elmalaki}

\begin{abstract}
Thanks to the rapid growth in wearable technologies, monitoring complex human context becomes feasible, paving the way to develop human-in-the-loop IoT systems that naturally evolve to adapt to the human and environment state autonomously. Nevertheless, a central challenge in designing such personalized IoT applications arises from human variability. Such variability stems from the fact that different humans exhibit different behaviors when interacting with IoT applications (intra-human variability), the same human may change the behavior over time when interacting with the same IoT application (inter-human variability), and human behavior may be affected by the behaviors of other people in the same environment (multi-human variability). To that end, we propose \textbf{FaiR-IoT}, a general reinforcement learning-based framework for adaptive and fairness-aware human-in-the-loop IoT applications. In FaiR-IoT, three levels of reinforcement learning agents interact to continuously learn human preferences and maximize the system's performance and fairness while taking into account the intra-, inter-, and multi-human variability. We validate the proposed framework on two applications, namely (i) Human-in-the-Loop Automotive Advanced Driver Assistance Systems and (ii) Human-in-the-Loop Smart House. Results obtained on these two applications validate the generality of FaiR-IoT and its ability to provide a personalized experience while enhancing the system's performance by $40\%-60\%$ compared to non-personalized systems and enhancing the fairness of the multi-human systems by 1.5 orders of magnitude.

\end{abstract}

\ccsdesc[500]{Computing methodologies~Multi-agent reinforcement learning}
\ccsdesc[500]{Human-centered computing~Ubiquitous computing}
\ccsdesc[300]{Human-centered computing~Collaborative and social computing}

\keywords{Fairness, Personalized IoT, Reinforcement Learning, Human Adaptation, Human-in-the-Loop}

\maketitle

\thispagestyle{empty}

\vspace{-2mm}
\section{Introduction}\vspace{-1mm}
Ubiquitous computing-that interacts and adapts to humans-is inevitable. In these pervasive systems, human reactions and behavior are observed and coupled into the loop of computation~\cite{elmalaki2015caredroid}. By allowing autonomy into the essence of IoT systems, these evolving systems provide services that are adaptable to the human context and intervene and take actions that are tailored to the human reaction and behavior. 
Despite the IoT system's ability to collect and analyze a significant amount of sensory data, traditional IoT typically depends on fixed policies and schedules to enhance user experience. However, fixed policies that do not account for variations in human mood, reactions, and expectations, fail to achieve the promised user experience. Moreover, with the continuous and inevitable interaction of the human with these systems, it becomes a pressing need to adapt to the physical environment changes and adapt to human preferences and behavior. This opens the question of how to use the monitored human state to design human-in-the-loop IoT systems that provide a personalized experience.

A personalized IoT system needs to ``infer'' or ``learn'' the human preference and continuously ``adapts'' and ``takes actions'' whether autonomously or in the form of recommendation. Hence, we need a \textbf{human-in-the-loop framework} that moves from a one-size-fits-all approach to a personalized process in which learning and adaptation agents in IoT systems are tailored towards humans' individual needs. This feedback property opens the door to design a ``Reinforcement Learning (RL)'' based agent. Unfortunately, applying standard RL algorithms, such as Q-learning, faces several challenges in the context of human-in-the-loop IoT applications. In particular, adapting to the human reaction and behavior poses a new set of challenges due to the \textbf{intra-human} and \textbf{inter-human} variability—for example, the same human behavior and reaction change over time. Even for a small period of time, the same human may produce different reactions based on unmodeled external effects. Similarly, different humans have different reactions under similar conditions. In addition to the variability introduced by humans, the IoT system used to infer human states suffers itself from variability (e.g., power constraints, connectivity status, classification/signal processing errors) that introduce another level of complexity. Moreover, as IoT applications are becoming more ubiquitous, multiple humans may interact within the same application space affecting each other's reaction and the way the application adapts. This \textbf{multi-human} interaction poses a new set of challenges to the IoT application relating to the ``fairness'' of adaptation.

In this paper, we propose a framework that can be used in conjunction with IoT applications to provide a personalized experience while addressing the aforementioned challenges. We purpose \textbf{Fairness-aware Human-in-the-Loop Reinforcement Learning framework}, or \textbf{FaiR-IoT} for short, that monitors the change in the human reaction and behavior while interacting with the IoT system and addresses the \textbf{intra-human} and \textbf{inter-human} variability, as well as, the \textbf{multi-human} interactions, to provide personalized adaptation that enhances the human's experience while ensuring fairness across all human sharing the same IoT application. The proposed framework is then used to build two human-in-the-loop IoT applications, (1) personalized advanced driver assistance system and (2) personalized smart home.
\subsection{Related Work} \vspace{-1mm}
\textbf{Human-in-the-Loop IoT:}
Besides the fact that IoT solutions nowadays have billions of connected devices~\cite{middleton2013forecast}, humans themselves are becoming a walking sensor network equipped with wearables that are rich in sensors and network capabilities~\cite{nunes2015survey, elmalaki2021neuro}. 
Human sensing has a central role in many IoT applications. In the area of managing the energy consumption of buildings, detecting the number of occupants through human sensing has been a primary target of many energy-saving techniques, such as correlating electrical load usage with occupancy sensors~\cite{balaji2013sentinel, rowe2011sensor}. 
In automotive applications, the human driving behavior has been integrated into the loop of computation of many Advanced Driver Assistance Systems (ADAS)~\cite{dua2019autorate, nambi2018hams}, such as activate the automatic cruise control~\cite{moon2008human} or adjust the threshold of the forward collision warning~\cite{elmalaki2018sentio}. In smart cities, human tracking is used for dynamic resource allocation of community services~\cite{tsai2017towards}. Moreover, as IoT applications are becoming more human-centric applications, efficient human modeling, human state estimation, and human adaptation are vital components for IoT applications that interact with humans~\cite{li2020deepalerts, wang2010markov}. 
In this paper, we propose a general framework that can be integrated into many of these applications to provide a personalized experience while addressing the human variability that arises while interacting with these applications.  

\textbf{Reinforcement learning for human adaptation:}
This tight coupling between human behavior and computing promises a radical change in human life~\cite{picard2000affective}. In the area of cognitive learning and human-in-the-loop IoT applications, reinforcement learning (RL) has proven to be adequate to monitor human intentions and responses~\cite{sadigh2016information, sadigh2017active, hadfield2016cooperative}. 
Multisample RL~\cite{elmalaki2018sentio} can adapt to humans and changes in their response times under various autonomous actions. Amazon has used personalized reinforcement learning to adapt to students' preferences~\cite{bassen2020reinforcement}. Moreover, RL models have been used to decide residential load scheduling~\cite{zhang2019cooperative}.
In this paper, we build upon work in RL literature to monitor changes in human behavior to achieve personalized IoT applications while adapting the RL models at runtime to address the different variability that arises in human-in-the-loop IoT systems. 

\textbf{Hierarchical reinforcement learning:}
Multi-layer RL has been used in the domain of parameter search~\cite{jomaa2019hyp}. In particular, RL has been used to learn policies to converge to good hyper-parameters that achieve the best performance~\cite{xu2017reinforcement}. 
In addition to parameter tuning, hierarchical RL was used to train multiple levels of policies or to decompose complex learning tasks into sub-goals~\cite{nachum2018data}. 
However, these methods learn the hyper-parameters once and fit them to the model or learns the different layers of policies and fix them. In the domain of human-in-the-loop IoT systems, this is not applicable. Humans change their behavior, and there exists intra-human and inter-human variability that needs to be addressed.

\textbf{Fairness in reinforcement learning:}
The question of fairness in RL where the agent prefers one action over another becomes more significant in multi-agent systems~\cite{jabbari2017fairness, joseph2016fairness}. As each agent learns its optimal policy, the notion of efficiency (system performance) and fairness may be conflicting. One approach is to train each agent independently but design the reward to be fair-efficient~\cite{jiang2019learning}. Another approach is the notion of envy and guilt in the study of inequity aversion to shape the fair reward function~\cite{hughes2018inequity}. However, in the multi-human application space, a policy that is considered fair at one time may become a discriminatory policy after some time as human preferences change. In this paper, we build upon the definition of fairness proposed in literature while proposing a fairness measure that can be embedded in human adaptation models.

\vspace{-1mm}
\subsection{Paper Contribution}\vspace{-1mm}
This paper aims to put the intrinsic variation in human behavior and reaction into the loop of computation while addressing the fairness in multi-human IoT applications. In particular, we make the following contributions:
\begin{enumerate}[leftmargin=*, noitemsep,topsep=0pt]
    \item Designing \textbf{\sysname}, an adaptation framework named ``Fairness-aware Human-in-the-Loop Reinforcement Learning'' that addresses the \textbf{intra-human} and \textbf{inter-human} variability in the context of IoT applications. The framework continuously monitors the human state through the IoT sensors and the changes in the environment with which the human interacts to adapt to the environment accordingly and enhance the human's experience. 
    \item Extending the framework to address the \textbf{multi-human} interaction within the same application space. The framework continuously monitors how each human behavior can affect other humans' behaviors, which changes the environment and the way it is adapting.    
    \item Proposing fairness-aware methodology for adapting to multiple humans sharing the same IoT application space. 
    \item We use the proposed framework to develop two Human-in-the-Loop IoT applications in the context of advanced driver assistance systems (ADAS) and automated smart home. 
\end{enumerate}

\vspace{-1mm}
\section{Fairness-aware Human-in-the-Loop Reinforcement Learning Framework}\vspace{-1mm}

\begin{figure}[!t]
\centering
    \includegraphics[width=0.99\columnwidth, trim={0.4cm 3.5cm 12cm 0.85cm},clip]{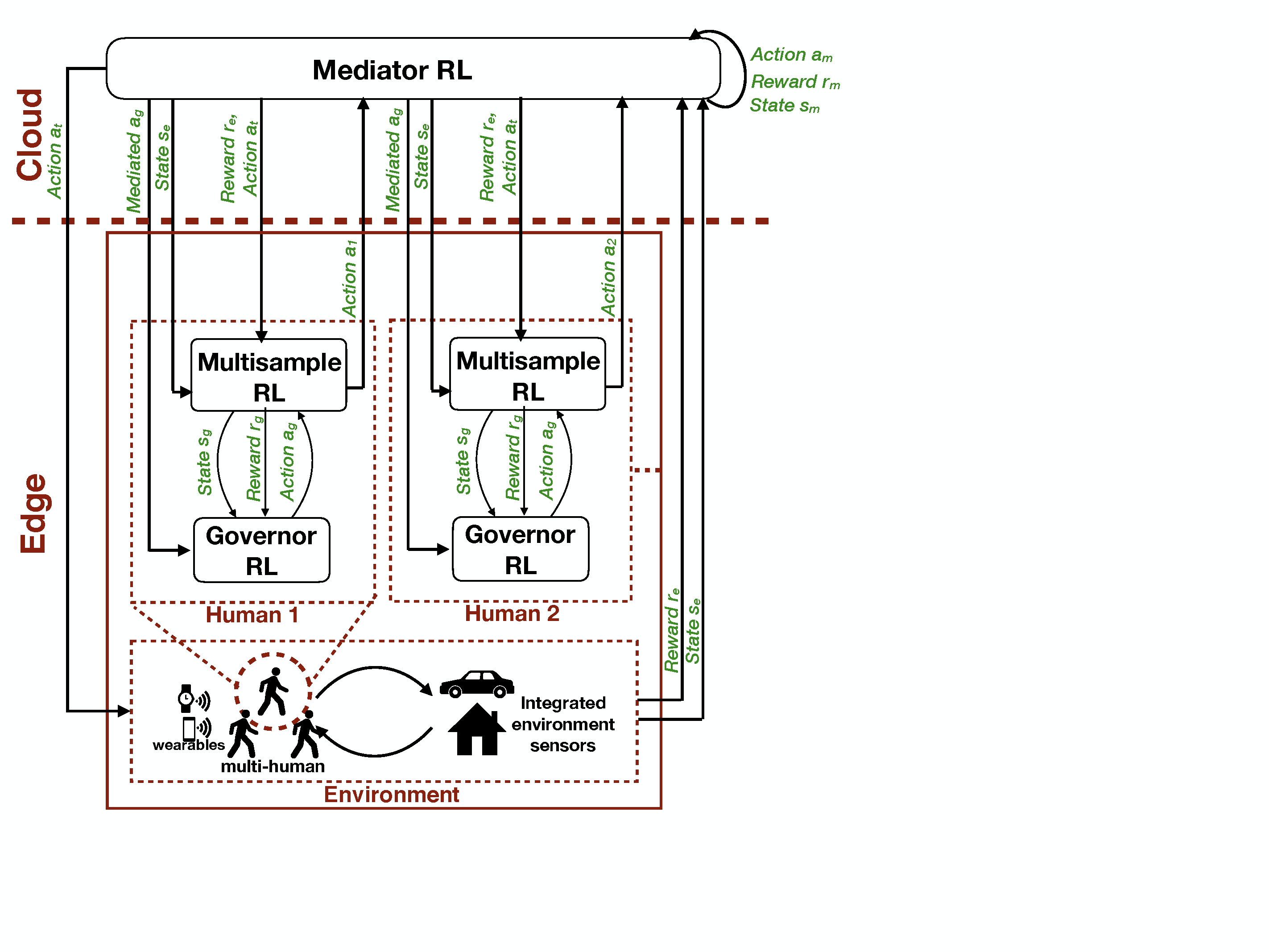}
    \caption{
       FaiR-IoT framework is divided into three RL agents. Multisample RL and Governor RL interact through action $a_g$, state $s_g$, and reward $r_g$ to handle the intra-human and inter-human variability at the edge. While a third RL agent, the Mediator RL, interacts with the environment through action $a_t$, state $s_e$, and reward $r_e$ in the cloud while providing a feedback action to both the Multisample RL and the Governor RL to handle the multi-human variability.
    } \label{fig:framework}
  \vspace{-6mm}
\end{figure}

We consider one of the IoT applications in the automated smart home as a motivating example for designing the proposed framework. In particular, we focus on the application of designing smart thermostats. 
Current state-of-art smart thermostats can adapt the home temperature based on room occupation~\cite{balaji2013sentinel} using fixed schedules and policies~\cite{lu2010smart}. In particular, preset configurations are provided by homeowners, and smart thermostat abides by these configurations. However, human needs and behavior vary across time, such as a change in the body's temperature is affected by multiple aspects, such as excitement, anxiety, physical activity, and health issues. In principle, IoT systems can play a role in detecting the human mood~\cite{adib2015smart} and activity~\cite{guo2017wifi} to ``adapt'' the home temperature accordingly. This adaptation can further be automated and ``tuned'' by monitoring the human thermal satisfaction, which can be inferred as direct feedback from the human~\cite{pritoni2017occupant} or as an indirect measurement from another edge device, such as, black globe thermometer~\cite{bedford1934globe} or skin temperature monitor~\cite{sim2018wearable}. The end goal is to achieve a personalized experience by learning the best home temperature that provides the best thermal sensation for the current human state. This ``learn'', ``adapt'', and  ``tune'' feedback loop fits well with the RL paradigm. 
The environment in the RL setting encapsulates the interaction between the human and the IoT applications. The RL agent makes a recommendation to the human or takes action (such as changing the set-point of the thermostat) to adapt the IoT application. This action or recommendation is continuously being adapted to match the changes in human behavior (such as mood and physical activity) and reaction (such as a change in the thermal sensation). This model keeps training online or offline until it converges to the best policy, which is the best thermostat set-point for a particular human mood and physical activity. However, we argue that this setting will not achieve the overarching goal of personalized experience in IoT applications. Using the same motivating example, we list the cases in which the setting mentioned above will fall short.
\begin{enumerate}[leftmargin=*, noitemsep,topsep=0pt]
    \item \textbf{Intra-human variability:} Even under the same mood and physical activity, the same human may change the personal preference for the best room temperature. Hence, an RL model that learns a fixed policy will not be adequate. 
    \item \textbf{Inter-human variability:} Different humans may have different thermal sensations even with the same home temperature under the same mood and physical activity. Moreover, the human body response to a change in ambient temperature can differ across humans~\cite{du2014response}. Hence, the same RL model design (same parameters and same reward function) may adapt correctly for some humans while performing poorly for others.
    \item \textbf{Multi-human variability:} More than one human can be in the same house, each with a different mood (relaxed or stressed) and different physical activity (sleeping or doing domestic work). One set-point that achieves the most comfortable thermal sensation for one human may not be the best one for another. Hence, an RL model that does not take the effect of the interaction between multiple humans within the same IoT application space will not satisfy the individual needs and will not account for the fairness of the adaptation policy across the humans. 
\end{enumerate}

Accordingly, we propose \textbf{``Human-in-the-Loop Fairness-aware Reinforcement Learning''} framework to address this variability. Our framework is divided into three components name \textbf{Multisample RL} to handle the intra-human variability, \textbf{Governor RL} to handle the inter-human variability, and \textbf{Mediator RL} to handle the multi-human variability while ensuring fairness. A conceptual figure for the proposed framework is shown in Figure~\ref{fig:framework}.  

\begin{table*}[!t]
\centering
     \begin{tabularx}{\textwidth} {p{0.55\textwidth}| p{0.42\textwidth} } 
     \includegraphics[trim={5cm 12cm 1cm 0.5cm},clip, width=0.55\textwidth]{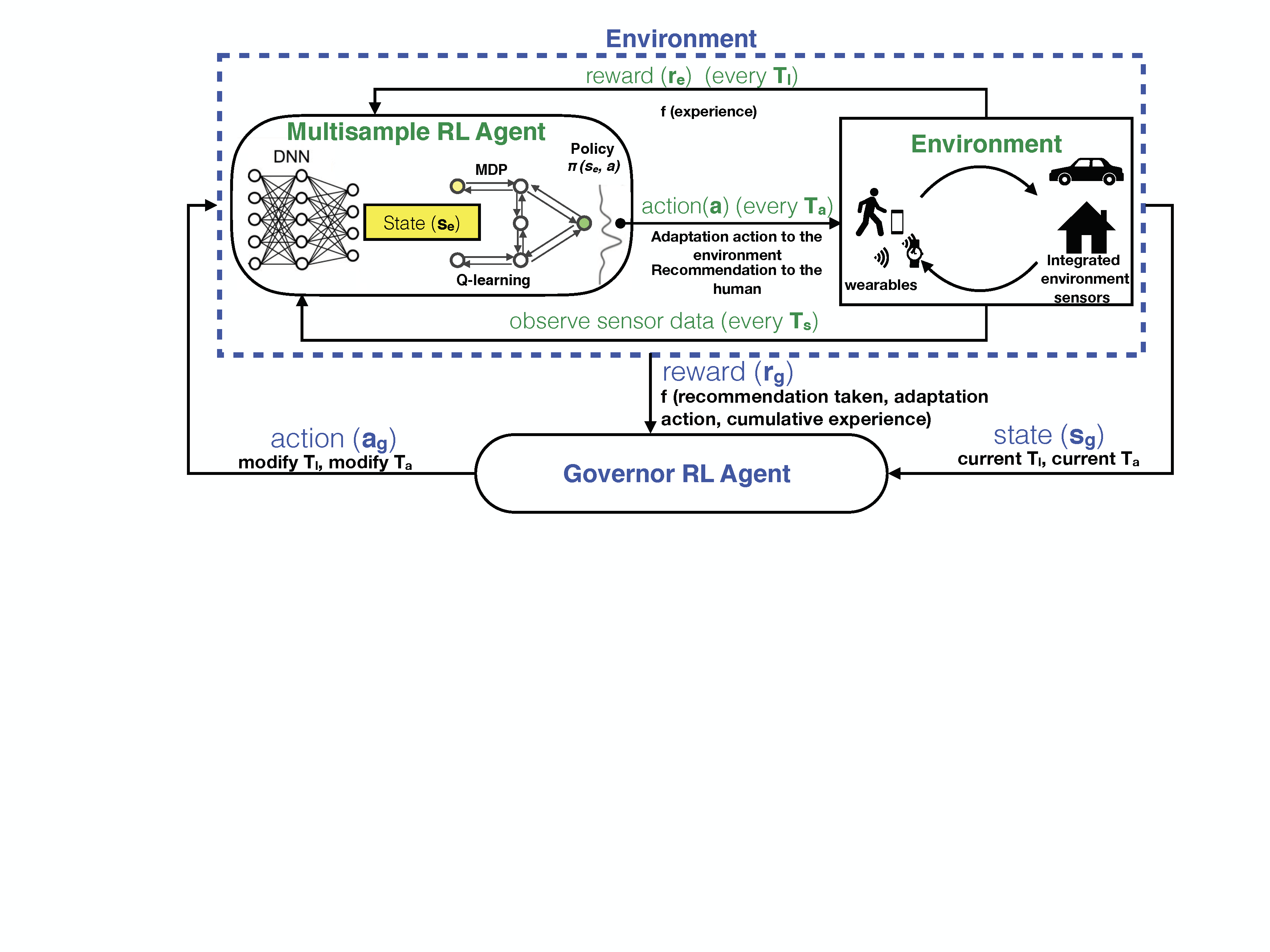}
     \captionof{figure}{Governor RL interacts with Multisample RL to adapt to the inter-human variability. Governor RL adapts the values of $T_l$ and $T_a$}\vspace{-3mm}
     \label{fig:governorRL}  & 
     \includegraphics[trim={0cm 13cm 13cm 0cm},clip, scale=0.35]{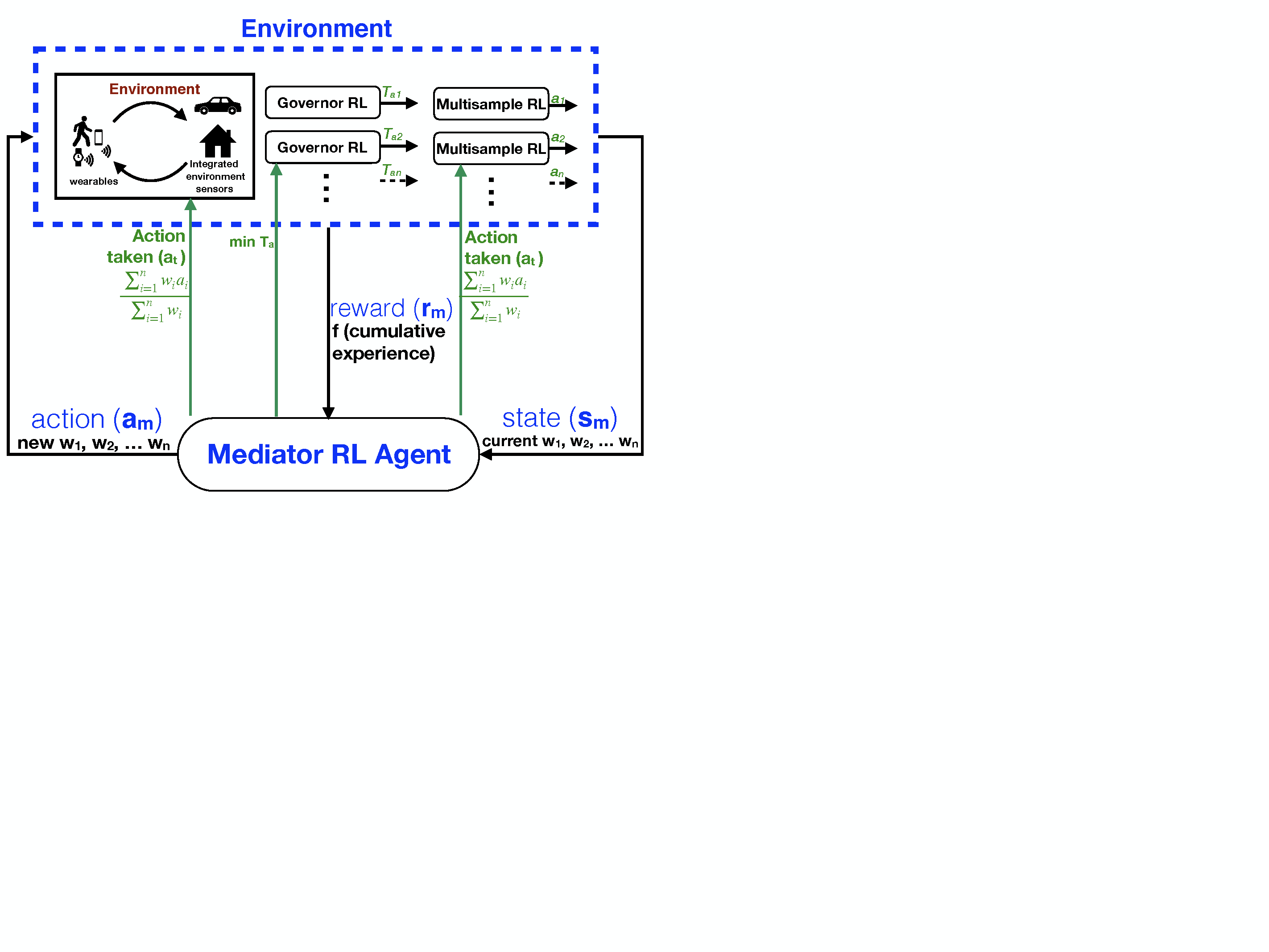}
     \captionof{figure}{Mediator RL adapts to multi-human variability by adapting the weights of the adaptation actions. 
     } \vspace{-3mm}
     \label{fig:mediatorRL}
\end{tabularx}
\vspace{-4mm}
\end{table*}

\vspace{-1mm}
\subsection{Intra-Human Variability}\label{sec:intra}\vspace{-1mm}
In the domain of human-in-the-loop IoT application, the changes in the human response time and the preferences across time makes the reward for the RL agent a time-varying function. The problem of time-varying reward function was addressed by \textbf{Multisample Q-learning algorithm}~\cite{elmalaki2018sentio} that divides the time horizon into three overlapping different time scales, named (1) state observing rate ($T_s$) where the state of the environment is observed every time new sensor data is available. This depends on the sampling rate of the sensors at the edge devices, (2) actuation rate ($T_a$), which is the time by which the agent should take action, and (3) evaluation rate ($T_l$) where the reward---corresponding to a particular taken action---should be evaluated at a relatively slow rate to take into account the delay in the environment change.

Multisample RL can handle the \textbf{intra-human variability}, where humans can change their behavior and reaction, and the continuous reward calculation tracks this change and adapts the RL action accordingly. However, it assumes that the response time of all humans is the same. By fixing the values of $T_l$ and $T_a$, Multisample RL assumes that all humans have the same response time. However, $T_l$ and $T_a$ should be adaptive based on the human interaction with the IoT system. By adapting $T_l$ and $T_a$, we can take the \textbf{inter-human variability} as well into the loop of computation. This leads to designing \textbf{Governor RL} as will be explained in the next section.

\vspace{-1mm}
\subsection{Inter-Human Variability}\label{sec:inter}\vspace{-1mm}
To account for the inter-human variability, the RL-agent needs to adapt $T_l$ and $T_a$ assumed by the Multisample RL algorithm to personalize the human experience. Such adaptation can be modeled as another RL-agent that observes the change in the human's experience as the parameters $T_l$ and $T_a$ change. To that end, and as shown in Figure~\ref{fig:governorRL}, we add a \textbf{Governor RL} layer that personalizes the values of $T_l$ and $T_a$ for Multisample RL. The Multisample RL underneath runs the Q-learning algorithm with the three overlapped time scales ($T_l$, $T_a$, and $T_s$) and adapts the IoT system that interacts with the human through an action or a recommendation to the human. A personalized Q-learning policy is learned based on the reward propagated from the environment. This reward is a function of the human experience and the IoT system performance. After the policy is converged (the reward value is not increasing), the Governor RL observes the cumulative reward for this particular action $a_g$ that modifies the values of $T_l$ and $T_a$ and adapts them accordingly. Eventually, Governor RL  will converge to the values of $T_l$ and $T_a$ that achieve the best cumulative reward for a particular human. The algorithm and the design of the Governor RL will be discussed in Section~\ref{sec:governorRL}.

\vspace{-2mm}
\subsection{Multi-Human Variability}\label{sec:multi}\vspace{-1mm}
When multiple humans interact in the same IoT application space, their preferences differ based on the number of people they interact with and the type of this interaction. Moreover, when the application adapts to one human, the environment changes accordingly to the adaptation action, and thus it has a direct effect on the other humans in the same environment. In addition to that, multiple humans may require different adaptation actions based on their different preferences. Hence, we need a mediator to intervene and manage these multiple adaptation actions from multiple humans to achieve an aggregate better state for all the humans collectively. Accordingly, we propose the third layer in our framework, the \textbf{Mediator RL}. As shown in Figure~\ref{fig:mediatorRL}, the Mediator RL takes the adaptation actions ($a_1$, $a_2$, $...$, $a_n$) from all the Multisample RL agents. 

As mentioned in Section~\ref{sec:intra}, $T_l$ is the evaluation rate of the RL agent, while $T_a$ is the actuation rate that governs the time to apply the action on the environment. Since $T_l$ captures the environment's response time, including the human response time, it should not change even while mediating multiple humans. However, the actuation rate $T_a$ should be mediated because it controls how often we apply an action to the environment. Hence, the Mediator RL has to mediate two values, (1) $a_g$:~which is the action taken by each Governor RL to adapt $T_a$, and (2) $a_i$: which is the action taken by each Multisample RL for each human. 

Accordingly, the Mediator RL is responsible for mediating $a_g$ by choosing the smallest value of $T_a$ across all $a_g$. This will ensure the fastest actuation rate across all humans. However, by changing the actuation rate $T_a$, Governor RL has to be notified accordingly because now the reward that the Governor RL observes is accounted for a different value of $T_a$. Hence, to ensure that the reward is associated with the right action, the Mediator RL echos back the mediated $a_g$, which has the same $T_l$ but a different $T_a$.

Next, the Mediator RL should choose the right action to apply to the environment. To that end, and not to compromise the effectiveness of the action on the human experience, we can not just choose one action from the list of actions. Hence, we use a weighted average of all the actions. In principle, the weighted average takes into account the varying degrees of importance of the numbers. However, since we do not know apriori, which action is more important than others to achieve a better cumulative experience for all humans, we can not fix these weights.
Moreover, as humans' preferences, moods, and reactions are changing, one action may have more weight at a particular time while having less weight at another time. Hence, we need to learn these weights and continuously adapt them based on all humans interacting in the same environment. Accordingly, the Mediator RL agent adjusts the actions' weights and then applies the weighted average action $a_t$ to the environment. The Mediator RL agent then observes the effect of these weights by collecting the reward $r_m$, a cumulative reward from all the humans' experiences in the environment. However, since the actual action taken on the environment is different from the individual actions $a_i$ by the different Multisample RL, $a_t$ has to be echoed back to every Multisample RL to associate the reward it is observing with the correct action.
Moreover, to ensure that the Mediator RL does not keep on favoring one action over the rest. We extend the design of the Mediator RL to include a fairness measure combined with the humans' experiences to calculate the reward $r_m$. The design of the Mediator RL will be discussed in Section~\ref{sec:algorithm}.

\section{Algorithm Design}\label{sec:algorithm}

This section describes how the Governor RL and the Mediator RL agents are designed and how they interact with each other. 

\subsection{Design of the Governor RL Agent}\label{sec:governorRL}
We model the problem of adapting the values of $T_l$ and $T_a$ as executing an optimal policy over a Markov Decision Process (MDP) whose states are defined over $T_l$ and $T_a$ and whose reward function captures the IoT system performance. By executing such an optimal policy, the RL agent is guaranteed to search over the space of $T_l$ and $T_a$ values to maximize the system's performance. 
The MDP main components namely state space, action space, and reward function are designed as follows: 
\vspace{-1mm}
\noindent \paragraph{\textbf{Governor MDP State space $\mathcal{X_G}$:}} The MDP states are the different values that $T_l$ and $T_a$ can take. That is, each state $s_g \in \mathcal{X_G}$ is determined by a tuple ($T_l$, $T_a$). Since RL agents' performance depends heavily on the cardinality of the state space, we discretize the continuous values of $T_l$ and $T_a$ into a finite number of values where such discretization depends on the context of the application as it will be discussed in Sections~\ref{sec:app1} and~\ref{sec:app2}.

\noindent\paragraph{\textbf{Governor MDP Action space $\mathcal{A_G}$:}}  The action space $\mathcal{A_G}$ in this MDP contains all the possible combinations of changing the $T_l$ and $T_a$ values. Particularly, the action can be either \textit{decrement} with some value $\downarrow$, \textit{increment} with some value $\uparrow$, or \textit{remain} the same value $\circlearrowleft$. Such actions are defined for both $T_l$ and $T_a$ under the constraint that $T_l\geq T_a$. We designed the actions as an increment or a decrement rather than choosing a specific tuple ($T_l$, $T_a$) from all the possible combinations to make sure that there is no big sudden change in the actuator rate ($T_a$) which may compromise the human's experience.

\vspace{-1mm}
\noindent\paragraph{\textbf{Governor MDP Reward Function $R_{\mathcal{G}}$:}}
Each state $s_g \in \mathcal{S_G}$ is associated with a performance $p_s$. Such performance $p_s$ is a measure of the human's experience and the performance of the IoT system, which depends on the context of the application, as it will be discussed in Section~\ref{sec:headerapps}.
The performance is calculated after running the Multisample $Q$-learning \cite{elmalaki2018sentio} using the values of $T_l$ and $T_a$ associated with state $s_g$. The associated performance $p_s$ plays a role in determining the reward $r_g$ value that accrued due to taking the action $a_g$ at the state $s_g$. In particular, the reward value $r_g = R_{\mathcal{G}}(s_g, a_g)$ (positively or negatively) depends on a weighted difference between the performance of the system $p_s$ at the state $s_g$ (before modifying the values of $T_l$ and $T_a$) and the performance of the system $p_{s'}$ at the new state $s'_g$ (after modifying the values of $T_l$ and $T_a$).

In this MDP setting, the transition probabilities are known apriori since \textit{increasing/decreasing/remain the same} $T_l$ or $T_a$ leads to a known state. However, the reward is unknown apriori because it depends on the human's experience. Moreover, it can change over time. Hence, to solve the MDP when the reward values are unknown, we use $Q$-learning, which is summarized in Algorithm~\ref{alg:q-learning-governor}.

\setlength{\textfloatsep}{1pt}
\begin{algorithm}[!t]
\small
        \begin{algorithmic}
        \State \textbf{Hyper parameters:} Learning parameters: $\alpha$, $\gamma$, $\epsilon$
        
        \State \textbf{Require:}
        \Statex Sates $\mathcal{X_G} = \{(T_l, T_a)_1, \dots, (T_l, T_a)_x\}$
        \Statex Actions $\mathcal{A_G} = \{(\circlearrowleft T_l, \circlearrowleft T_a), (\circlearrowleft T_l, \uparrow T_a), (\uparrow T_l, \circlearrowleft T_a), (\uparrow T_l, \uparrow T_a), $
        \Statex $\qquad\qquad\quad$ $(\circlearrowleft T_l, \downarrow T_a), (\downarrow T_l, \circlearrowleft T_a), (\downarrow T_l, \downarrow T_a), (\downarrow T_l, \uparrow T_a),$
        \Statex $\qquad\qquad\quad$ $(\uparrow T_l, \downarrow T_a) \}$, 
        \Statex Reward function $R_{\mathcal{G}}: \mathcal{X_G} \times \mathcal{A_G} \rightarrow \mathbb{R}$
        \Statex Transition function $T_{\mathcal{G}}: \mathcal{X_G} \times \mathcal{A_G} \rightarrow \mathcal{X_G}$
        \Statex Multisample $Q$-learning algorithm: $MuQL(T_l,T_a)$
        \Statex Learning rate $\alpha \in [0, 1]$,  $\alpha = 0.9$
        \Statex Discounting factor $\gamma \in [0, 1]$, $\gamma = 0.1$
        \Statex $\epsilon$-Greedy exploration strategy $\epsilon \in [0, 1]$, $\epsilon = 0.2$
        \Statex Weighted Performance Difference $\mathcal{W}$
        \Procedure{$GovQL$}{$\mathcal{X_G}$, $\mathcal{A_G}$, $R_{\mathcal{G}}$, $T_{\mathcal{G}}$, $\alpha$, $\gamma$}
            \State Initialize $Q_{\mathcal{G}}: \mathcal{X_G} \times \mathcal{A_G} \rightarrow \mathbb{R}$  with 0
            \While{true}
                \State Start in state $s_g \in \mathcal{X_G}$
                \State $p_s \gets MuQL(s_g)$ \Comment{Calculate the performance of $s_g$}
                
                    \State Calculate $\pi(s_g)$ according to $Q_{\mathcal{G}}$ and exploration strategy:
                    \State\qquad with probability $\epsilon$: $\pi(s_g) \gets$ choose an action a at random,
                    \State\qquad with probability $1-\epsilon$: $\pi(s_g) \gets \argmax_{a} Q_{\mathcal{G}}(s_g, a)$)
                    \State $a_g \gets \pi(s_g)$
                    \State $s'_g \gets T_{\mathcal{G}}(s_g, a_g)$ \Comment{Receive the new state}
                    \State $p_{s'} \gets MuQL(s'_g)$ \Comment{Calculate the performance of $s'_g$}
                    \State $r_g \gets R_{\mathcal{G}}(s_g, a_g) = \mathcal{W}(p_{s'},p_s)$ \Comment{Receive the reward}
                    \State $Q_{\mathcal{G}}(s'_g, a_g) \gets (1 - \alpha) \cdot Q_{\mathcal{G}}(s_g, a_g) + \alpha \cdot (r_g + \gamma \cdot \max_{a'} Q_{\mathcal{G}}(s'_g, a'))$
                    \State $s_g \gets s'_g$~\\
                
            \EndWhile
            \Return $Q_{\mathcal{G}}$
        \EndProcedure
        \end{algorithmic}
    \caption{Governor RL} 
    \label{alg:q-learning-governor}
\end{algorithm}

\subsection{Design of the Mediator RL Agent}\label{sec:mediatorRL}
We model the problem of learning the best weights ($w_1$, $w_2$, ... $w_n$) for the adaptation actions collected for individual personalized performance in a multi-human setting ($a_1$, $a_2$, ..., $a_n$) as executing an optimal policy over a Markov Decision Process (MDP). 

\vspace{-1mm}
\noindent\paragraph{\textbf{Mediator MDP State space $\mathcal{X_M}$:}}
The MDP states are the different values of $w_i$ that can be summed up to $1$. We discretize their continuous values into finite number of state space $\mathcal{X_M}$, where their values are chosen from a predefined set of values $\{0, 0.2, 0.4, 0.6, 0.8, 1\}$. The state space's size depends on the number of actions that we need to take their weighted average, reflecting the number of humans in the environment. 

\vspace{-1mm}
\noindent\paragraph{\textbf{Mediator MDP Action space $\mathcal{A_M}$:}}
The action space $\mathcal{A_M}$ of the Mediator MDP contains all the possible jumps $\nearrow$ to all the states with the same probability. Such a design choice allows the Mediator RL to rapidly switch between weights and converge faster to the optimal assignment of weights.

\noindent\paragraph{\textbf{Mediator MDP Reward function $R_\mathcal{M}$:}}
Each state $s_m$ has a performance $p_s$ that is associated with it. The performance $p$ is a measure of all the humans' cumulative experience and the performance of the IoT system, which depends on the context of the application, as will be discussed in Section~\ref{sec:app2}. 
The performance is calculated after applying the weighted average action ($a_t$ = $\ \frac{\sum_{i=1}^n w_i a_i}{\sum_{i=1}^n w_i}$) at a rate of the minimum $T_a$ across all individual humans (as chosen by their individual Governor RL). Each human will have a different experience with the applied action $a_t$. Hence, the cumulative performance $p_s$ associated with this state $s_m$ will be a function of all the human experiences, which depends on the context of the application. 
To calculate the reward, we use the same idea of calculating the reward in Governor RL. In particular, the associated reward value $r_m$ is computed by the reward function $R_{\mathcal{M}}$($s_m$,$a_m$) which is a function of the relative performance between two states $s_m$ (which corresponds to the current weights) and the next state $s'_m$ (which corresponds to the new weights). 
In this MDP setting, the reward is unknown apriori as it depends on the cumulative humans' experiences. Moreover, it can change over time. Hence, to solve the MDP when the reward values are unknown, we use the $Q$-learning summarized in Algorithm~\ref{alg:q-learning-mediator}. In particular, the algorithm is divided into two procedures, $MEDQL$, which runs the Q-learning algorithm in which the reward value depends on the performance of the current state $s_m$ and next state $s'_m$. The performance of each state is calculated in another procedure $CalculateStatePerformance$ that takes a state $s_m$ as input and gets the preferred actions per human (by running $MuQL$) and the different actuation rates (by running $GovQL$) to calculate the cumulative performance $p_s$.

\setlength{\textfloatsep}{1pt}
\begin{algorithm}[!h]
\small
        \begin{algorithmic}
        \State \textbf{Hyper parameters:} Learning parameters: $\alpha$, $\gamma$, $\epsilon$
        
        \State \textbf{Require:}
        \Statex Humans $\mathcal{H_M} = \{h_1, h_2, \dots, h_n\}$
        \Statex Sates $\mathcal{X_M} = \{(w_1, w_2, \dots, w_n)_1, \dots, (w_1, w_2, \dots, w_n)_x\}$
        \Statex Actions $\mathcal{A_M} = \{(\nearrow 1, \nearrow 2, \dots,  \nearrow x) \}$ 
        \Statex Reward function $R_\mathcal{M}:           \mathcal{X_M} \times \mathcal{A_M} \rightarrow \mathbb{R}$
        \Statex Transition function $T_\mathcal{M}: \mathcal{X_M} \times \mathcal{A_M} \rightarrow \mathcal{X_M}$
        \Statex Multisample $Q$-learning algorithm per human $h$: $MuQL_{h}$ 
        \Statex Governor $Q$-learning algorithm per human $h$: $GovQL_{h}$
        \Statex Weighted Average function: WAVG (l, x) = $\ \frac{\sum_{i=1}^n l(i)x(i)}{\sum_{i=1}^n x(i)}$
        \Statex Learning rate $\alpha \in [0, 1]$,  $\alpha = 0.9$
        \Statex Discounting factor $\gamma \in [0, 1]$, $\gamma = 0.1$
        \Statex $\epsilon$-Greedy exploration strategy $\epsilon \in [0, 1]$, $\epsilon = 0.2$
        \Statex Weighted Performance Difference $\mathcal{W}$
        
        \Procedure{$CalculateStatePerformance$}{$s_m$}
            \State Actions Array: $Act=\{\}$
            \State Actuation Rates Array: $ARates=\{\}$
            \State Human Experience Array: $HExp_{s_m}=\{\}$ 
            \State $\forall h \in \mathcal{H_M}$, $Act$.insert($a_h$), $s.t.$ $a_h \gets$ current action by $MuQL_h$ 
            \State $\forall h \in \mathcal{H_M}$, $ARates$.insert($T_a$), $s.t.$ $T_a \gets$ current $T_a$ by $GovQL_h$ 
            \State $a_{t_{s_m}} \gets $ WAVG(Act, $s_m$)
            \State $min T_a \gets \min\{ARates\}$
            \State $\forall h \in \mathcal{H_M}$, Update $MuQL_h$ with $a_{t_{s_m}}$
            \State $\forall h \in \mathcal{H_M}$, Update $GovQL$ with $min\ T_a$
            \State $\forall h \in \mathcal{H_M}$, $HExp_{s_m}$.insert($r_e$), 
            \State \qquad \qquad \qquad \qquad $s.t.$ $r_e \gets$ CalculatePerformance $(h, a_{t_{s_m}})$ 
            \State $p_s \gets $CalculateCumulativePerformance($HExp_{s_m}$) 
            
            \Return $p_s$
        \EndProcedure
        
        \Procedure{$MedQL$}{$\mathcal{X_M}$, $\mathcal{A_M}$, $R_\mathcal{M}$, $T_\mathcal{M}$, $\alpha$, $\gamma$}
            \State Initialize $Q_\mathcal{M}: \mathcal{X_M} \times \mathcal{A_M} \rightarrow \mathbb{R}$  with 0
            \While{true}
                \State Start in state $s_m \in \mathcal{X_M}$
                \State $p_s \gets CalculateStatePerformance(s_m)$ 
                \State \Comment{Calculate the performance of $s_m$} 
                    \State Calculate $\pi$ according to $Q_{\mathcal{M}}$ and exploration strategy:
                    \State\qquad with probability $\epsilon$: $\pi(s_m) \gets$ choose an action a at random,
                    \State\qquad with probability $1-\epsilon$: $\pi(s_m) \gets \argmax_{a} Q_\mathcal{M}(s_m, a)$)
                    \State $a_m \gets \pi(s_m)$
                    \State $s'_m \gets T_{\mathcal{M}}(s_m, a_m)$ \Comment{Receive the new state}
                    \State $p_{s'} \gets CalculateStatePerformance(s'_m)$ 
                    \State \Comment{Calculate the performance of $s'_m$}
                    \State $r_m \gets R_\mathcal{M}(s_m, a_m) = \mathcal{W}(p_{s'},p_s)$ \Comment{Receive the reward}
                    \State $Q_{\mathcal{M}}(s'_m, a_m)\gets(1-\alpha)\cdot Q{\mathcal{M}}(s_m, a_m) $ 
                    \State \qquad\qquad \qquad\qquad$+ \alpha \cdot (r_m+\gamma\cdot\max_{a'} Q_{\mathcal{M}}(s_m', a'))$
                    \State $s_m \gets s'_m$
            \EndWhile
            \Return $Q_{\mathcal{M}}$
        \EndProcedure
        \end{algorithmic}
    \caption{Mediator RL} 
    \label{alg:q-learning-mediator}
\end{algorithm}

\subsection{Design of Fairness-aware Mediator RL}\label{sec:fair-mediator}\vspace{-1mm}
The Mediator RL tries to maximize the total reward by choosing different weights for the different adaptation actions learned from multiple humans preferences. In one extreme case, as we will show in the results (Section~\ref{fig:exp4}), the Mediator RL may end up favoring one adaptation action over the others (by keeping assigning high weight to it) to increase its total reward. This means that the Mediator RL favors one human preference over the others because it looks at the cumulative performance. To model this inequality, we borrow concepts from sociology which named this phenomenon the ``Matthew effect'', which is summarized as the rich get richer and the poor get poorer~\cite{bol2018matthew}. Accordingly, to ensure fairness across all humans, we need to avoid the ``Matthew effect'' by modifying the reward function of the Mediator RL. In particular, we model the Mediator RL as a resource distributed among multiple humans, and we want to ensure the fair-share of this resource. This leads us to use the notion of utility. 
We define the utility of each human adaptation $h$ at timestep $t$ as:
\begin{align*}
u_{h_t} = \frac{1}{t} \sum_{j=0}^{t} \frac{j}{t} w_{h_j} .
\end{align*}

In particular, $u_{h_t}$ is the average weight assigned by the Mediator RL for a particular human $h$ over a time horizon $[0:t]$, where the factor $\frac{j}{t}$ is used to give more value to the recent weights learnt by the Mediator RL over the ones in the past. We then measure the fairness of the Mediator RL using the coefficient of variation ($cv$) of the human utilities~\cite{jain1984quantitative}: \[ cv = \sqrt{\frac{1}{n-1} \sum_{h=1}^{n} \frac{(u_h - \bar{u})^2}{\bar{u}^2}},\] where $\bar{u}$ is the average utility of all humans. The Mediator RL is said to be more fair if and only if the $cv$ is smaller. Accordingly, we modify the reward function $R_\mathcal{M}$ to reflect the changes in $cv$. In particular, in Algorithm~\ref{alg:q-learning-mediator}, we add a new fairness measure $\mathcal{F}$ which denotes the difference between the $cv$ values at $s_m$ and $s'_m$. Hence, we change $r_m$ to become: \[ r_m \gets R_\mathcal{M}(s_m, a_m) = (1-\zeta) \mathcal{W}(p_{s'},p_s) + \zeta \mathcal{F}(cv_{s'}, cv_s),\] where $\zeta$ is used as a fairness scale from $0$ to $1$. As we will show in the results (Section~\ref{fig:exp4}), the objective to increase the performance measure $\mathcal{W}$ and the objective to increase the fairness measure $\mathcal{F}$ can be two opposing objectives. 

\section{Implementation and Evaluation Plan}\label{sec:headerapps}
We applied the proposed framework to two applications, the first one is in the domain of Advanced Driver Assistance System (ADAS), and in particular, we focus on the Forward Collision Warning (FCW) application. The second application is in the smart home automation system domain, and in particular, we focus on heating, ventilation, and air conditioning systems (HVAC). To ensure the safety of the real human subjects, we decided to use a simulation environment for both applications\footnote{At the time of preparing this manuscript, the COVID-19 pandemic was hindering against conducting experiments with real human subjects. Social distancing and isolation were enforced in the authors' location.}. In each application, we will explain the used simulator, the various human behaviors, how the performance is calculated, and how the proposed \sysname framework can track the intra-human, inter-human, and multi-human variability.

\begin{figure*}[!t]
\centering
     \begin{tabularx}{\textwidth} {p{0.31\textwidth}| p{0.31\textwidth} | p{0.31\textwidth}} 
     \multicolumn{1}{c|}{\textbf{Human $H_1$}} & 
     \multicolumn{1}{c|}{\textbf{Human $H_2$}} & 
     \multicolumn{1}{c}{\textbf{Human $H_3$}}\\ \hline
     \includegraphics[width=0.31\textwidth]{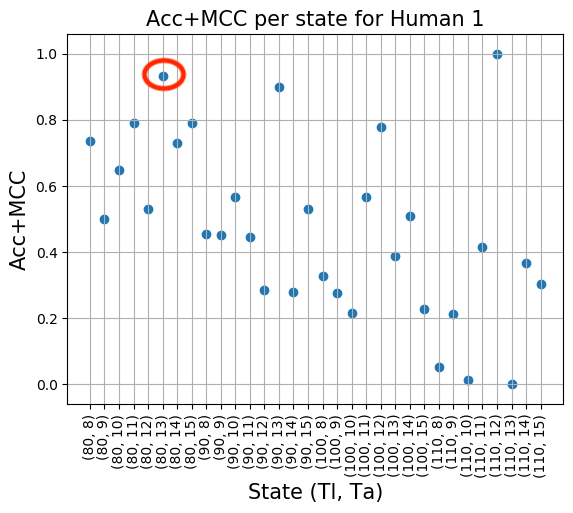}
     & 
     \includegraphics[width=0.31\textwidth]{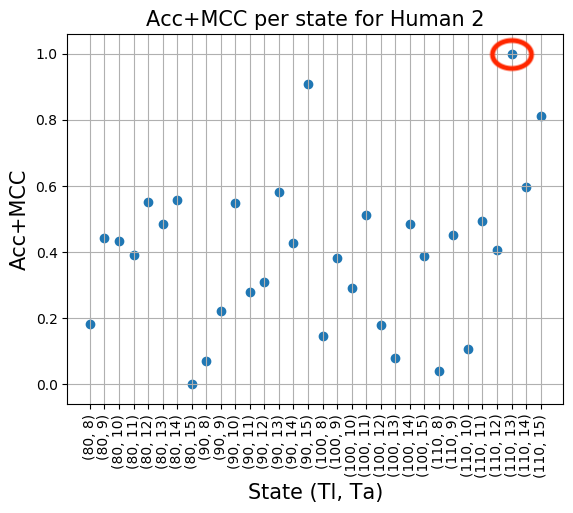}
     &
     \includegraphics[width=0.31\textwidth]{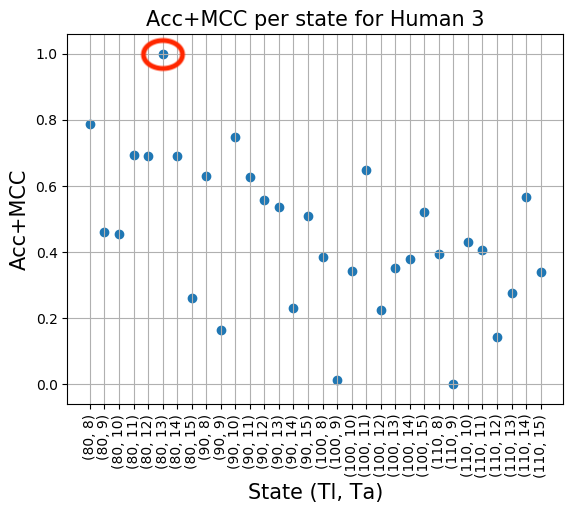}
     \\
     \includegraphics[width=0.31\textwidth]{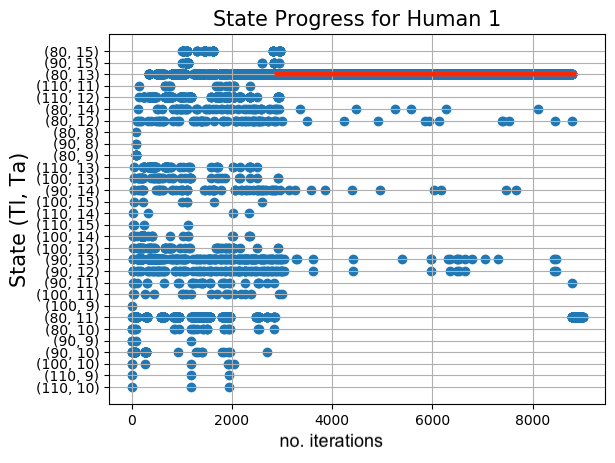}
     &
     \includegraphics[width=0.31\textwidth]{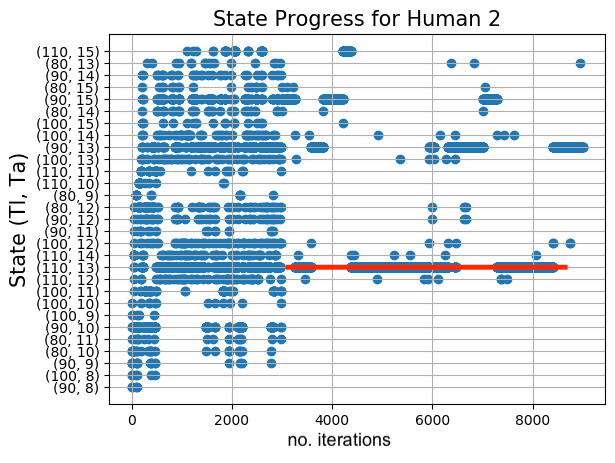}
     &
     \includegraphics[width=0.31\textwidth]{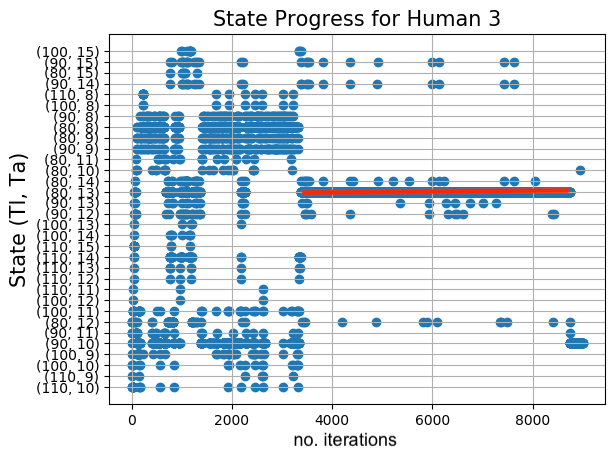}
     \\
     \includegraphics[width=0.31\textwidth]{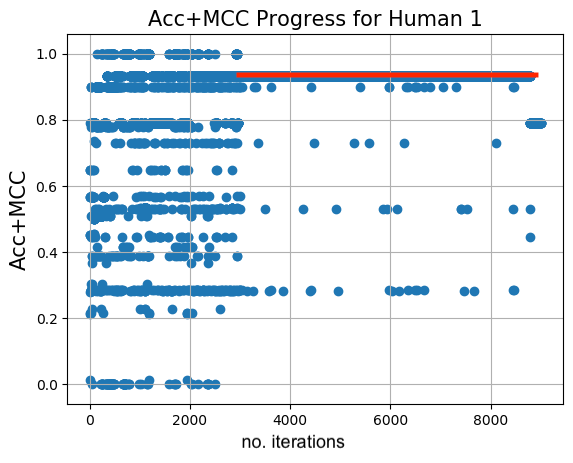}
     & 
     \includegraphics[width=0.31\textwidth]{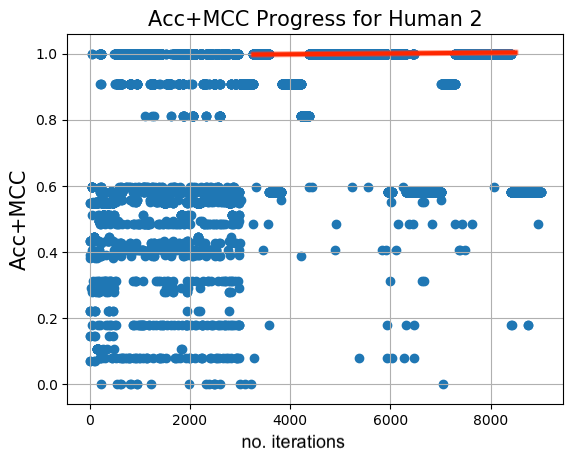}
     & 
     \includegraphics[width=0.31\textwidth]{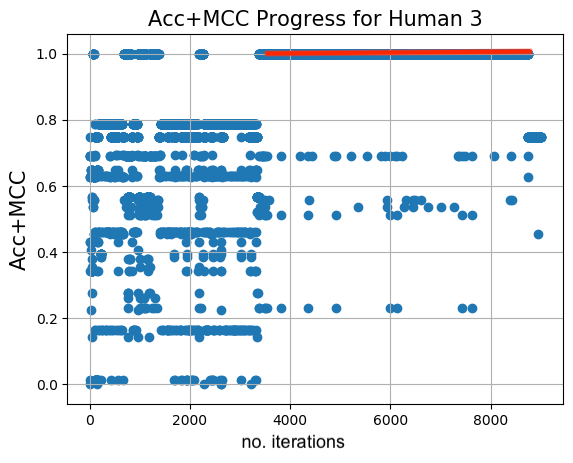}
    \end{tabularx}
\caption{Numerical results of Experiment 1. (Top) the performance of the system---obtained through brute force search---under different assignments for $T_L$ and $T_a$ for humans $H_1$ on the left, human $H_2$ in the middle, and human $H_3$ on the right. (Center) the progress of the states of the Governor RL agent across different iterations of the system showing the convergence to the states with maximum system performance. (Bottom) the performance of the system due to the adaptations of the Governor RL agent showing an increase in the performance over time.} 
\label{fig:exp1}
\vspace{-3mm}
\end{figure*}

\section{Application 1: Human-in-the-Loop ADAS - Forward Collision Warning (FCW)}\label{sec:app1}
Standard FCW system measures the time-to-crash based on the distance and the relative velocity of the front object and, if the time-to-crash is below a certain threshold (signaling a possible risk of collision), it alerts the driver to apply the brakes. A human-in-the-loop FCW should take the human state and preferences while calculating this threshold. For example, if the driver is distracted, the alarm should be displayed earlier. Moreover, some drivers take more time to react to the alarm and press the brakes (response time). Hence, the alarm threshold should count for this response time as well. Personalized FCW has been addressed in the literature using offline learning of the threshold based on statistical modeling to reduce the required data~\cite{Muehlfeld2013statistical}, drivers' expected response deceleration (ERDs)~\cite{wang2016development}, parameter identification of driver behavior is proposed in~\cite{wang2016forward}, and adapting the time of the FCW based on the response time and the mental state of the driver~\cite{elmalaki2018sentio}.

\vspace{-1mm}
\noindent \paragraph{\textbf{Simulator:}}
We used the Skoda Octavia model~\cite{octaviamodel} for vehicle dynamics to simulate the physics of both the driving car and the leading car. We interfaced the vehicle model with the Matlab virtual reality toolbox (VR) to virtualize the vehicle dynamics. Each vehicle has parameters that specify its mass, the wheels' mechanical friction on the ground, its acceleration, and braking forces. 
\vspace{-1mm}
\noindent \paragraph{\textbf{Simulated human subjects:}}
Following the observations in~\cite{elmalaki2018sentio} to model driving behavior, we simulated three humans with three different driving behavior as follows: 
\begin{itemize}[leftmargin=*,noitemsep,topsep=0pt]
    \item \textbf{Moderate driver behavior $H_1$}: This driver has a moderate behavior with average braking intensity, acceleration intensity, average relative distance to the leading car, and average response time. 
    \item \textbf{Aggressive driver behavior $H_2$}: This driver tends to apply high intensity on both the brakes and the acceleration pedal. This is accompanied by a small relative distance to the leading car and a short response time. 
    \item \textbf{Slow driver behavior $H_3$}: This driver tends to be more conservative. The driver tends to apply low intensity of both the brakes and the acceleration pedal. This is accompanied by a large relative distance to the leading car and a larger response time.
\end{itemize}

\vspace{-1mm}
\noindent \paragraph{\textbf{Adapting to the intra-human and inter-human variability using Governor RL-agent:}}
As described in Algorithm~\ref{alg:q-learning-governor}, the state of MDP is a tuple of ($T_l$, $T_a$). The response time of the driver and the behavior of the driver are unknown apriori. Different values of $T_l$ and $T_a$ may result in better performance and better driving experience for different humans (inter-human variability). We quantize the state space into $32$ states. The sampling time $T_s$ is fixed at $0.25$ seconds while $T_l$ and $T_a$ are multiples of the sampling time. In particular, we choose $T_l \in [80, 90, 100, 110]$ and $T_a \in [8, 9, 10, 11, 12 ,13, 14, 15]$ multiples of $T_s$. A state $s_g$ is one combination of $T_l$ and $T_a$, such as (80,8). The different states capture the inter-human variability where different humans can have different response times which entail different learning rate ($T_l$) and different actuation rate ($T_a$).

\vspace{-1.5mm}
\noindent\paragraph{\textbf{Performance function:}}
The performance is measured in terms of the following parameters: 
\begin{itemize}[leftmargin=*, noitemsep,topsep=0pt]
    \item False Negatives (FN): The relative distance between the leading car and the driving car is below the safety distance of 7m, which is calculated based on the 2 seconds rule~\cite{twoSecondsRule}. 
    \item False Positives (FP): The alarm with signaled unnecessarily. 
    \item True Positives (TP): The alarm is signaled, and the driver pressed the brakes. 
    \item True Negatives (TN): The alarm is not signaled, and the relative distance between the leading car and the driving car is above the safety distance of 7m.
\end{itemize}

Based on these parameters, the accuracy (Acc) and Matthews correlation coefficient (MCC) are then calculated for a specific state $s_g$. We combine both metrics (the Acc and the MCC) to measure the performance of state $s_g$. MCC has shown to be a good metric for binary classification evaluation~\cite{chicco2020advantages}, while accuracy gives the proportion of correct predictions. Hence, the performance $p_s$ as indicated in Algorithm~\ref{alg:q-learning-governor} is calculated as $p_s = Acc_s + MCC_s$. This value is normalized from 0 to 1.

\begin{figure*}[!t]
\centering
     \begin{tabularx}{\textwidth} {p{0.33\textwidth}| p{0.33\textwidth} | p{0.33\textwidth}} 
     \multicolumn{1}{c|}{\textbf{Switching from $H_1$ to $H_2$}} & 
     \multicolumn{1}{c|}{\textbf{Switching from $H_2$ to $H_1$}} & 
     \multicolumn{1}{c}{\textbf{Switching from $H_1$ to $H_3$}}\\ \hline
     \includegraphics[width=0.31\textwidth]{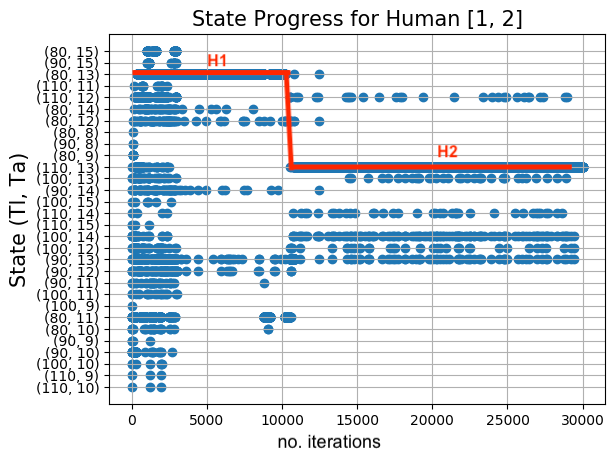}
     & 
     \includegraphics[width=0.31\textwidth]{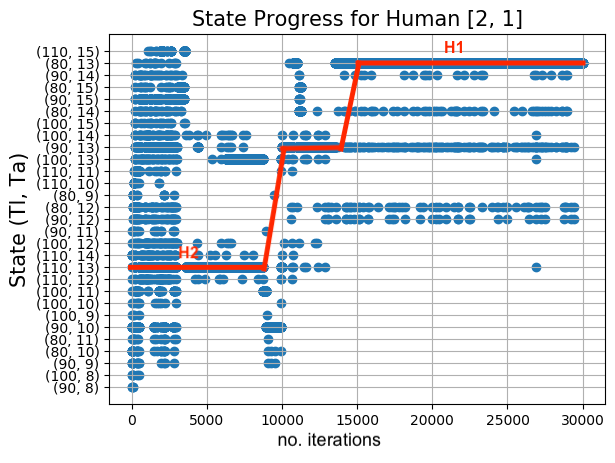}
     &
     \includegraphics[width=0.31\textwidth]{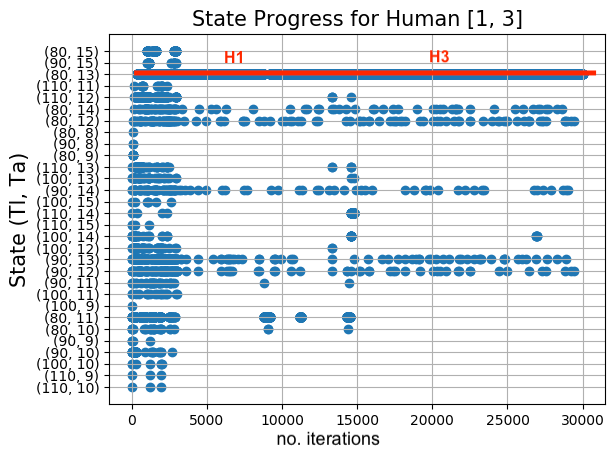}
     \\
     \includegraphics[width=0.31\textwidth]{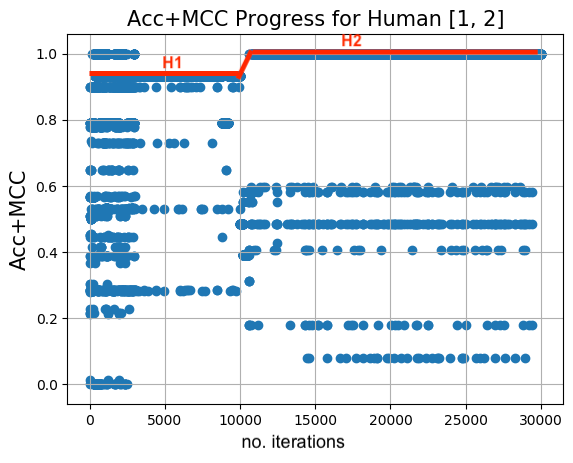}
     &
     \includegraphics[width=0.31\textwidth]{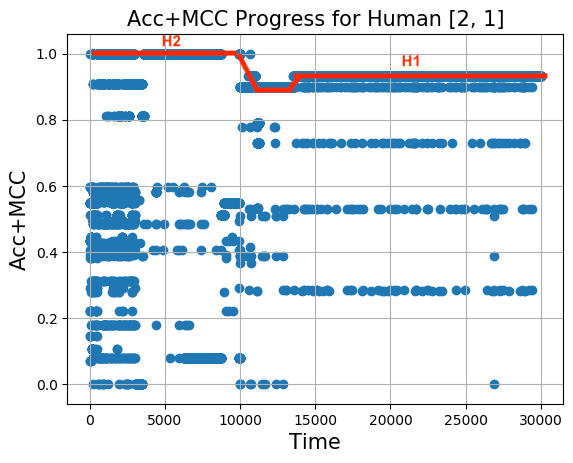}
     &
     \includegraphics[width=0.31\textwidth]{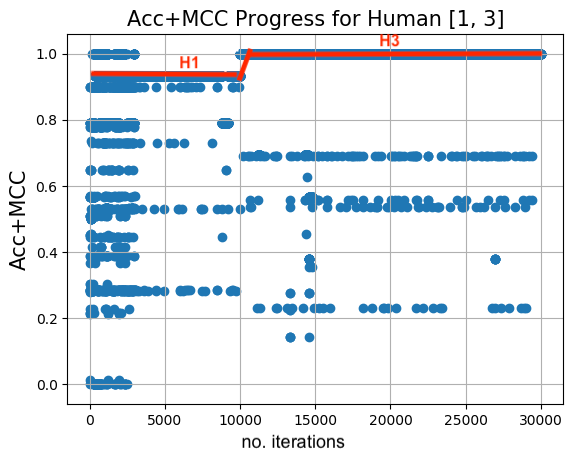}
    \end{tabularx}
\caption{Numerical results of Experiment 2. (Top) the progress of the states of the Governor RL across different iterations of the system when the simulated human behavior changes across time, showing the ability of the Governor RL agent to track states with maximum system performance. (Bottom) the performance of the system when the simulated human behavior changes across time, showing the ability of the Governor RL to maximize the system performance over time.}   
\label{fig:exp2}
\end{figure*}

\vspace{-1mm}
\subsection{Experiment 1: Inter-Human Variability}\vspace{-1mm}
The performance $p_s$ for each state $s_g$ indicated by ($T_l$, $T_a$) for humans $H_1$, $H_2$, and $H_3$ are calculated ahead of time to obtain the ground truth that allows us to validate the ability of Algorithm~\ref{alg:q-learning-governor} to obtain the values of ($T_l$, $T_a$) that maximizes the performance of the system. However, when the algorithm runs, it does not know apriori the performance value of a state $s_g$ until Multisample $Q$-learning runs for some time and returns the MCC and the Acc for this specific state.  
As shown in Figure~\ref{fig:exp1} (top), the best state for $H_1$ is (110,12) then (80,13) with a very small difference. Similarly, the best state for $H_2$ is (110,13), which was the worst for $H_1$, followed by (90,15). The best state for human $H_3$ is (80,13), then (80,8). These different states highlight the inter-human variability that needs to be accounted for to provide a personalized experience.

We run Algorithm~\ref{alg:q-learning-governor} for 9000 iterations. As shown in Figure~\ref{fig:exp1} (center), the algorithm converges to states (80,13), (110,13), and (80,13) for humans $H_1, H_2$ and $H_3$, respectively. These states are the second best state for $H_1$, and are the states with best performance for humans $H_2$ and $H_3$. This is also reflected in how the performance values $p_s$ (which is used to calculate the reward changes with the number of iterations for $H_1$, as shown in Figure~\ref{fig:exp1}(bottom) where the performance value converges to value $0.95$ which corresponds to state (80,13). As for $H_2$, the performance value converges to $1$, which corresponds to state (110,13), and for $H_3$ the performance value converges to $1$ which corresponds to state (80,13), as shown in Figure~\ref{fig:exp1} (bottom). 
These results show the proposed Governor RL's ability to track the inter-human variability and converge to the best (for $H_2$ and $H_3$) or near best (for $H_1$) performance for each human with different driving behavior.

\vspace{-1mm}
\subsection{Experiment 2: Intra-Human Variability}\vspace{-1mm}
We now study the proposed Governor RL's ability to adapt to the changes in human preference across time. To that end, to
show how the algorithm can trace the change in behavior to adapt to the intra-human variability. We switch between humans on the same simulator after 10000 iterations and observe how the algorithm will adapt to the new behavior. First, we switch from $H_1$ to $H_2$, then vice versa in Figure~\ref{fig:exp2} (top) and the respective performance progress across the number of iterations are shown in Figure~\ref{fig:exp2} (bottom). We repeat the same experiment but between $H_1$ and $H_3$ as seen in Figure~\ref{fig:exp2} and the respective performance progress across the number of iterations in Figure~\ref{fig:exp2}.
These figures show the ability of Governor RL to adapt to the changes in human behavior.

\section{Application 2: Human-in-the-Loop Smart House - A Thermal System}
Recent work in literature targets human-in-the-loop HVAC~\cite{jung2017towards} while trying to assist human satisfaction and addressing fairness across all occupants~\cite{shin2017exploring}. Reinforcement learning has been proposed to adapt the HVAC set-point based on human activity~\cite{elmalaki2018internet}. A human-in-the-loop thermal system should take the human state and preferences into the computation loop while calculating the heater set-point. For example, the human body temperature decreases when the human goes to sleep~\cite{barrett1993sleep}, while the body temperature increases when human exercises~\cite{lim2008human} and with stress and anxiety~\cite{olivier2003stress}. Monitoring the human state~\cite{likamwa2013moodscope}, sleep cycle~\cite{nguyen2016lightweight}, physical activity~\cite{shany2012sensors} are all possible using IoT edge devices.

\begin{table*}[!th]
\centering
\resizebox{.99\textwidth}{!}{
\begin{tabular}{|p{0.5cm} || p{0.5cm} | p{0.5cm} | p{0.5cm} | p{0.5cm} | p{0.5cm} | p{0.5cm} | p{0.5cm} | p{0.5cm} | p{0.5cm} | p{0.5cm} | p{0.5cm} | p{0.5cm} | p{0.5cm} | p{0.5cm} | p{0.5cm} | p{0.5cm} | p{0.5cm} | p{0.5cm} | p{0.5cm} | p{0.5cm} | p{0.5cm} | p{0.5cm} | p{0.5cm} | p{0.5cm} |}
 \hline
Hr & 0 & 1 & 2 & 3 & 4 & 5 & 6 & 7 & 8 & 9 & 10 & 11 & 12 & 13 & 14 & 15 & 16 & 17 & 18 & 19 & 20 & 21 & 22 & 23  \\\hline
 
 \textbf{$H_1$} & \multicolumn{6}{|c|}{$\star$} & $\star$ $\circledast$ $\spadesuit$ & $\Delta$ & \multicolumn{9}{|c|}{$\varphi$} & $\Delta$ $\bullet$ $\clubsuit$ & $\bullet$ $\clubsuit$ $\varheartsuit$ & $\bullet$ $\clubsuit$ $\varheartsuit$ $\vardiamondsuit$ & $\bullet$ $\circledast$ & $\bullet$ &\multicolumn{2}{|c|}{$\star$}\\ \hline

 \textbf{$H_2$} & \multicolumn{8}{|c|}{$\star$} & \multicolumn{2}{|c|}{$\star$ $\circledast$}& \multicolumn{9}{|c|}{$\varphi$} & \multicolumn{2}{|c|}{$\Delta$ $\bullet$ $\clubsuit$} & \multicolumn{2}{|c|}{$\bullet$ $\clubsuit$ $\varheartsuit$} & $\bullet$ $\circledast$ \\ \hline

\textbf{$H_3$} & \multicolumn{10}{|c|}{$\star$}  & $\star$ $\circledast$ & \multicolumn{4}{|c|}{$\varphi$} & $\Delta$ $\clubsuit$ & \multicolumn{2}{|c|}{$\bullet$ $\clubsuit$ $\vardiamondsuit$} & $\clubsuit$ $\varheartsuit$ & $\Delta$ $\clubsuit$ & \multicolumn{4}{|c|}{$\bullet$ $\circledast$ $\Delta$} \\ \hline
\end{tabular}
}
\caption{Human activity across the 24 hours of the day. Sleeping:$\star$, seated relaxed:$\bullet$, standing at rest:$\circledast$, standing light activity:$\Delta$, light domestic work:$\vardiamondsuit$, standing medium activity:$\clubsuit$, washing dishes standing:$\varheartsuit$, running:$\spadesuit$, and not home:$\varphi$. When multiple activities located in the same time slot, one of them is chosen randomly.}\label{tbl:app2behavior}
\vspace{-6mm}
\end{table*}

\vspace{-2mm}
\noindent\paragraph{\textbf{Simulator:}}
We simulated a mathematical model for a thermal house. In particular, we utilize a thermodynamic model of the house that considers the geometry of the house, the number of windows, the roof pitch angle, and the type of insulation used. The house is being heated by a heater with an airflow with a temperature of $50^{\circ}c$. A thermostat is used to allow fluctuation of $2.5^{\circ}c$ above and below the desired set-point, which specifies the temperature that must be maintained indoors. The set-point is controlled by an external controller that runs the proposed Governor RL algorithm. The heat flow into the house is calculated by:
\begin{align*}
\frac{dQ_{heater}}{dt} = (T_{heater} - T_{room}) \cdot \dot{M} \cdot c, 
\end{align*}

where $\frac{dQ}{dt}$ is the heat flow from the heater into the house, $c$ is the heat capacity of the air at constant pressure, and $\dot{M}$ is the air mass flow rate through the heater (set at $1 kg/sec$). $T_{heater}$ and $T_{room}$ are the temperature of hot air from the heater and the current room air temperature, respectively.

The thermal system in the house considers the heat flow from the heater and the heat loss to the environment. Heat losses to the environment depend on the thermal resistance of the house calculated by:
\vspace{-2mm}
\begin{align*}
    \frac{dQ_{losses}}{dt} = \frac{T_{room}-T_{out}}{R_{eq}},  
\end{align*}

where $T_{out}$ is the external temperature and $R_{eq}$ is the thermal equivalence of the house taken into consideration the thermal resistance of the wall and the windows. We assume that it is the winter season, so the $T_{out}$ is kept at $50^{\circ}F$ with a random range of daily temperature variation.

We model the humans as a heat source with heat flow that depends on the average exhale breath temperature (EBT) and the respiratory minute volume (RMV) ($\frac{dQ_{human}}{dt} \propto RMV \cdot EBT$)~\cite{gupta2010characterizing}.

The respiratory minute volume is the product of the breathing frequency ($f$) and the volume of gas exchanged during the breathing cycle, which are highly dependent on human activity. For example, $RMV \approx 6$ $l$/$m$ when the human is resting while $RMV \approx 12$ $l$/$m$ represents a human performing moderate exercise~\cite{carrollpulmonary2007}. We assume that the age/sex/time-of-day have no significance in the model. 

Finally, the indoor temperature-time derivative $\frac{dT_{room}}{dt}$ is directly proportional to the difference between the heat flow rate from the heater and the heat losses to the environment while taking into consideration the heat flow from the occupants. We calculate it as follows: 
\vspace{-3mm}
\begin{align*}
    \frac{dT_{room}}{dt} \propto\frac{dQ_{heater}}{dt} - \frac{dQ_{losses}}{dt} + \sum_{n=i}^{N} \frac{dQ_{human_{i}}}{dt},
\end{align*}
where $N$ is the number of occupants in the house. We simulated five days of a working human using this simulation environment.

\vspace{-1mm}
\noindent\paragraph{\textbf{Simulated human subjects:}}
We simulated the behavior of three humans based on their activity and their stress level.
The activity is simulated by the value of the RMV~\cite{carrollpulmonary2007} and the metabolic rate~\cite{metactivity}. Human stress level is simulated by an increase in the metabolic rate~\cite{olivier2003stress}.
We simulated nine activities arranged in the ascending order of RMV: sleeping, seated relaxed, standing at rest, standing light activity, light domestic work, standing medium activity, washing dishes standing, and running on a treadmill. We randomize the behavior by having different choices of activities in the same time slot. For example, as seen in Table~\ref{tbl:app2behavior},  $H_1$ can be doing one of three activities (sleeping, standing at rest, running) between 6 am to 7 am. In total, we simulated five working days for each human.       
\begin{itemize}[leftmargin=*, noitemsep,topsep=0pt]
    \item Occupant 1 ($H_1$) is an active human. $H_1$ wakes up early and exercises for one hour before leaving for work. When $H_1$ returns home, the activity changes according to the hour of the day.
    
    \item Occupant 2 ($H_2$) is a less active human. $H_2$ wakes up later than $H_1$ and does not perform any physical exercise before leaving for work. 
    
    \item Occupant 3 ($H_3$) is a less active human with more time at home.  
\end{itemize}

\begin{figure*}[!t]
\centering
     \begin{tabularx}{\textwidth} {>{\raggedright\arraybackslash}X  p{0.3\textwidth} | p{0.3\textwidth} | p{0.3\textwidth}} 
     \raisebox{1.0\height}{\rotatebox{90}{\textbf{Human 1}}}
     &
     \includegraphics[width=0.26\textwidth]{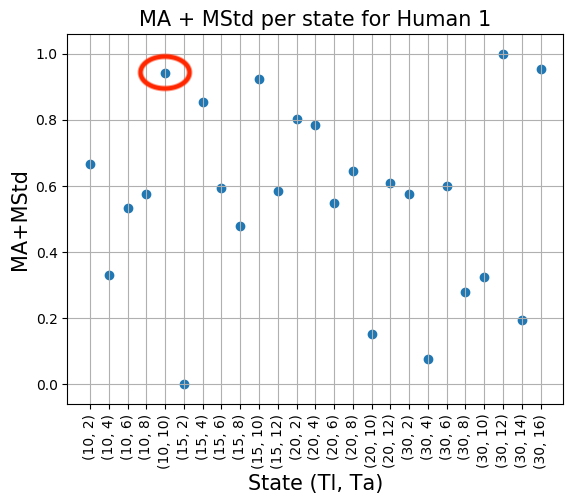}
     & 
     \includegraphics[width=0.3\textwidth]{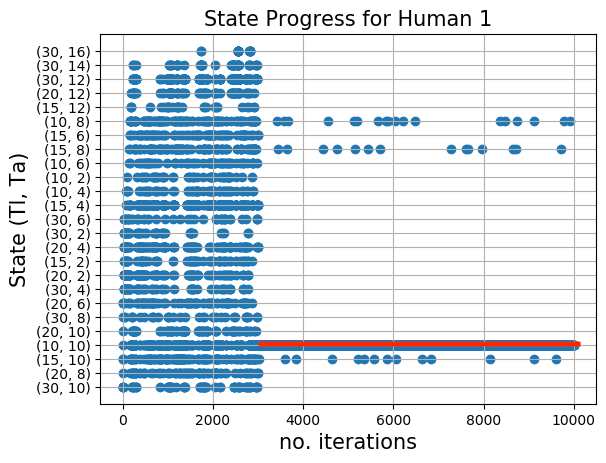}
     &
     \includegraphics[width=0.29\textwidth]{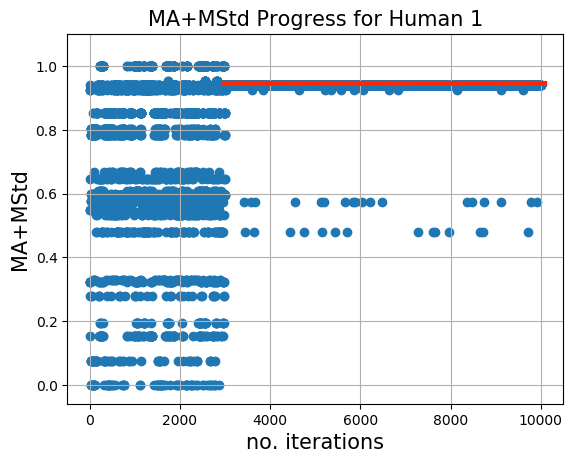}
     \\ 
     \raisebox{1.0\height}{\rotatebox{90}{\textbf{Human 2}}} 
     &
     \includegraphics[width=0.26\textwidth]{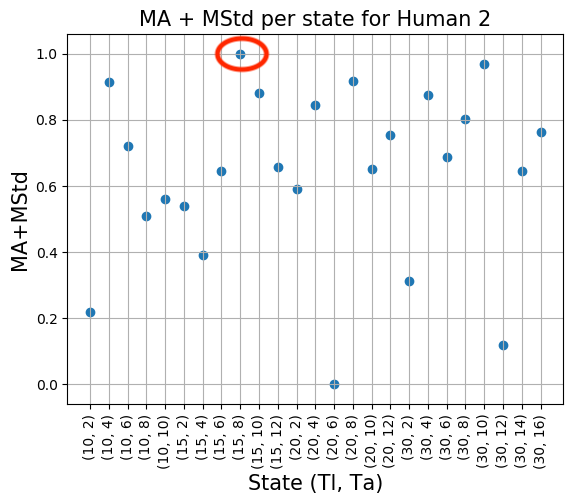}
     &
     \includegraphics[width=0.3\textwidth]{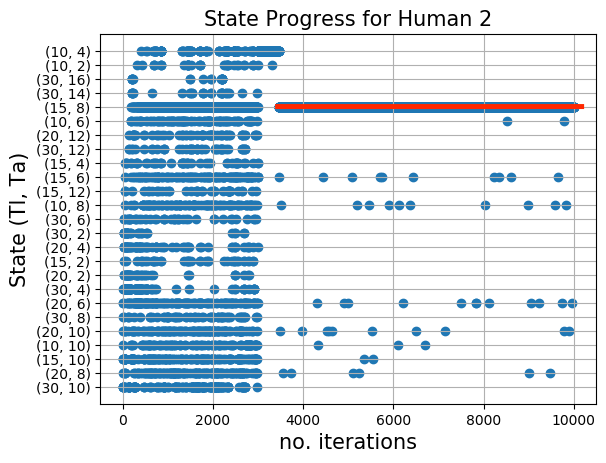}
     &
     \includegraphics[width=0.29\textwidth]{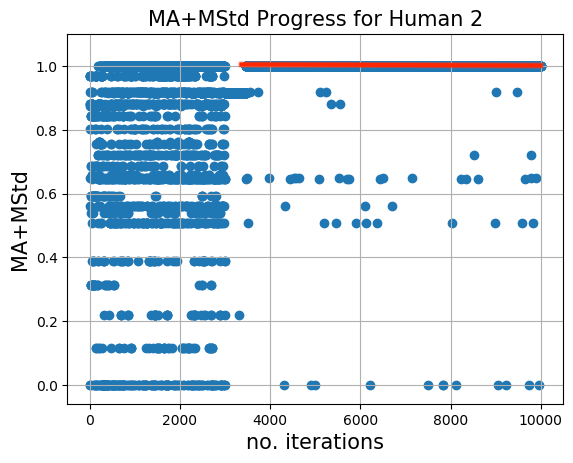}
    \vspace{-5mm}
    \end{tabularx}
\caption{Numerical results of Experiment 3. (Left) The system's performance-- obtained through brute force search-- under different assignments for $ T_L $ and $T_a$ for both human $H_1$ on the top and human $H_2$ on the bottom. (Center) the progress of the states of the Governor RL agent across different iterations of the system showing the convergence to the states with maximum system performance. (Right) The system's performance due to the adaptations of the Governor RL agent showing an increase in the performance over time.}
\label{fig:exp3}
\vspace{-3mm}
\end{figure*}

\noindent\paragraph{\textbf{Adapting to the intra-human and inter-human variability using Governor RL agent}:}
As described in Algorithm~\ref{alg:q-learning-governor}, the state of the MDP is a tuple of ($T_l$, $T_a$). This state is passed to Multisample $Q$-learning as a parameter. The thermal sensation, the behavior, and the state of the human are unknown apriori to the algorithm. Different values of ($T_l$) and ($T_a$) may give better thermal comfort for different humans (inter-human variability). We quantize the state space into $25$ states. The sampling time is $T_s$~=~$6$~min. $T_l$ and $T_a$ are multiples of the sampling time. In particular, $T_l \in [10, 15, 20, 30]$ and $T_a \in [2, 4, 6, 8, 10, 12, 14, 16]$ multiples of $T_s$ such that, $T_l \geq T_a$. A state $s$ is one combination of $T_l$ and $T_a$, such as $(15, 10)$. 
After choosing the MDP state for the Governor RL, Multisample $Q$-learning runs for simulated five days. We design the Multisample RL agent by modeling the human as an MDP with states corresponding to different activity ($Act$) and indoor temperature. 
A state $s$ is defined by a tuple of $T_{in}$ and $Act$. We quantize the indoor temperature ($T_{in}$) between $60^{\circ}$F to $80^{\circ}$F with $1^{\circ}$F granularity. When applying an action $a$ on a state $s$, it can transition to any other state with unknown transition probability due to the change in human behavior. The action space contains the different set-points between $70^{\circ}$F to $80^{\circ}$F with $2^{\circ}$F granularity.

The reward function $R(s, a)$ depends on the thermal comfort of the human, which can be estimated using Prediction Mean Vote (PMV)~\cite{fanger1970thermal}. PMV score indicates the thermal sensation of a human. It depends mainly on human activity, metabolic rate, clothing, and other environmental variables (airspeed, air temperature, mean radiant temperature, and vapor pressure). 
The scale of PMV ranges from $-3$ (very cold) to $3$ (very hot). According to ISO standard ASHRAE 55~\cite{handbook2009american}, a PMV in the range of $-0.5$ and $+0.5$ for the interior space is recommended to achieve thermal comfort. Estimation of the PMV score is calculated based on the knowledge of clothing insulation, the metabolic rate, the air vapor pressure, the air temperature, and the mean radiant temperature~\cite{fanger1970thermal}. The human thermal sensation can change for the same state due to unmodelled external factors. Hence, the reward value $r$ can change over time.
Moreover, each human can have a different response time (the time the human takes to feel a difference in thermal sensation). In addition to the variability introduced by the human, the response time of the thermal system (the time taken by the HVAC to actually reach the set-point) can be different and change with time due to system aging or unmodelled effects. Hence, $T_l$ and $T_a$ cannot be fixed and should differ from one human to another. 

\vspace{-1.5mm}
\noindent\paragraph{\textbf{Adapting to the multi-human variability using Mediator RL agent:}}
Since we modeled the human as a heat source, human activity affects the individual thermal sensation and affects indoor temperature. Moreover, when multiple humans exist in the same space and their activities are different, their thermal comfort will be at a different indoor temperature. To that end, to adapt to multiple humans in the same house, we need to measure their individual thermal comfort and then take their aggregate thermal comfort to find the HVAC set-point that can achieve a compromised thermal comfort for all the occupants. Accordingly, the HVAC set-point is controlled by running Algorithm~\ref{alg:q-learning-mediator} to mediate the diverse preferences of set-points from multiple humans. As mentioned in Algorithm~\ref{alg:q-learning-mediator}, a Governor RL with Multisample RL gets the individual preferred set-point per human ($a_t$), as well as the individual actuation rate $T_a$. 
States of the Mediator RL models the different weights for three set-points with a total of $21$ states. At each state $s_m$, Mediator RL takes the weighted average of the three set-points with the respective weights $a_{t_{s_m}}$ and sends it to the HVAC. For example, if the preferred set-points from the 3 humans are $72$, $75$, $78$, respectively, and the state $s_m$ is $(0.2, 0.6, 0.4)$, then $a_{t_{s_m}} = 75.5$, and hence the set-point that is sent to the HVAC is $76$ which is the nearest non-decimal number.

\vspace{-1mm}
\noindent\paragraph{\textbf{Performance function:}}
The performance is calculated based on two parameters, (1) the moving average of the absolute value of the PMV score over five days and (2) the moving standard deviation of the PMV score. This is used to account for high fluctuation in thermal sensation. The performance value depends on a weighted sum of these parameters. In particular, the performance $p_s$ for a particular state $s$ is calculated as follows: 
\begin{align*}
\overline{PMV}_\text{MA$_s$} &=  \frac 1n \sum_{i=0}^{n-1} \left| pmv_{-i}\right|, \qquad\qquad 
\overline{\sigma}_\text{MA$_s$} = \frac 1n \sum_{i=0}^{n-1} \sigma_{-i} \nonumber,\\
p_s &= \theta_1 \overline{PMV}_\text{MA$_s$} + \theta_2 \overline{\sigma}_\text{MA$_s$},
\label{eq:hvac_perf}
\end{align*}
where $\theta_1>\theta_2$ and $n$ is the number of samples. Indeed, the lower the value of the performance, the worst the thermal sensation that the human experiences.

\noindent\paragraph{\textbf{Reward functions:}}
As described in Algorithm~\ref{alg:q-learning-governor}, the reward function is calculated based on the performance of the current state $s_g$ (denoted as $p_s$) and the next state $s'_g$ (denoted as $p_{s'}$) after applying an action $a_g$. If $s'_g$ gives better performance than $s_g$, then it is a positive reward, otherwise, it is a negative reward. We use weighted performance difference between $p_s$ and $p_{s'}$ to calculate the reward value. The reward value $r_g$ is calculated according the following reward function $R_{\mathcal{G}}(s_g, a_g)$: 
\vspace{-1mm}
\begin{align*}
    \Delta &= | p_{s'} - p_s | \\
    \alpha &= \mathcal{W}(\Delta) \text{ , where $\mathcal{W}$ is an increasing step function.} \\
    R_{\mathcal{G}}(s_g, a_g) &= Sign(p_{s'} - p_s) \times \alpha \times p_{s'}.
\end{align*}

\subsection{Experiment 3: Inter-Human Variability}\vspace{-1mm}
We start by studying the proposed Governor RL agent's ability to adapt to individual human behaviors to select the optimal $(T_l, T_a)$ for each human. To that end, we apply the the same steps performed in application 1 in Section~\ref{sec:app1} for $H_1$ and $H_2$. Multisample $Q$-learning returns $\overline{PMV}_\text{MA$_s$}$ and $\overline{\sigma}_\text{MA$_s$}$ for a specific state $s_g$. By brute-forcing all combinations to discover the state with the maximum performance, we conclude that the best state for $H_1$ is (30,12) followed by (10,10) while the best states for $H_2$ are (15,8) and (30,10), as shown in Figure~\ref{fig:exp3} (left). It is worth noting that the best state for $H_2$ results in poor performance for $H_1$, highlighting again the inter-human variability that motivates the use of the proposed Governor RL.
We run Algorithm~\ref{alg:q-learning-governor} report the states chosen by the Governor RL across time in Figure~\ref{fig:exp3} (center)\footnote{Due to space limitation, we are showing the results for $H_1$ and $H_2$ only.}. We run the algorithm for a total of 10000 iterations with adaptive $\epsilon$ that decreases when the reward value does not change for multiple consecutive iterations. The algorithm converges to the states (10,10) and (15,8) for humans $H_1$ and $H_2$, respectively, which are the second-best and best states for the simulated humans. 
These results validate the designed Governor RL's ability to generalize to multiple applications and address the inter-human variability challenge.

\begin{figure*}[!th]
     \begin{tabularx}{\textwidth} {>{\raggedright\arraybackslash}X  p{0.2\textwidth} | p{0.25\textwidth} | p{0.22\textwidth} | p{0.22\textwidth}} 
    \raisebox{-0.3\totalheight}{\rotatebox{90}{\textbf{Without fairness}}}
     &
    \begin{subfigure}{0.2\textwidth}
    \raisebox{-0.3\totalheight}{\includegraphics[width=\textwidth]{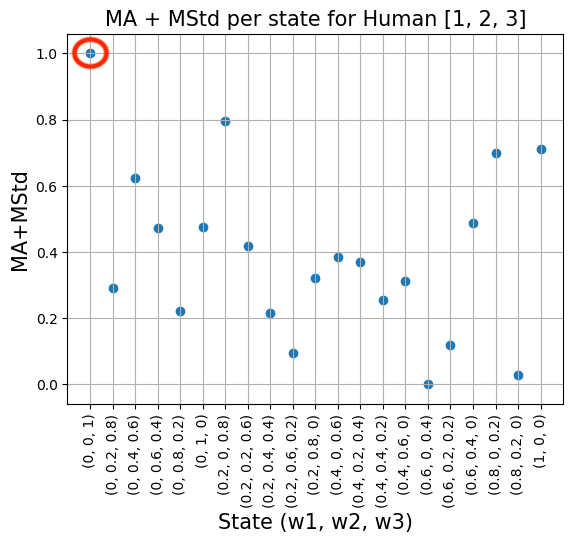}}\vspace{-1mm}
    \caption{}
    \label{fig:nocvmamstd} 
    \end{subfigure}
     & 
    \begin{subfigure}{0.25\textwidth}
    \raisebox{-0.3\totalheight}{\includegraphics[width=\textwidth]{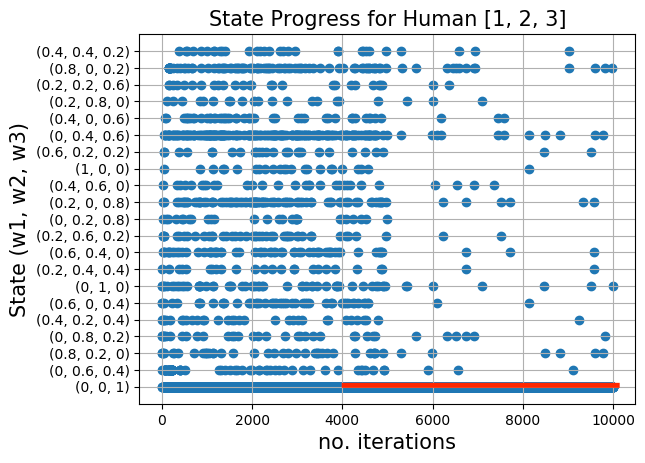}}
    \caption{}
    \label{fig:nocvstateprogressthermal} 
    \end{subfigure}
     &
    \begin{subfigure}{0.22\textwidth} 
    \raisebox{-0.3\totalheight}{\includegraphics[width=\textwidth]{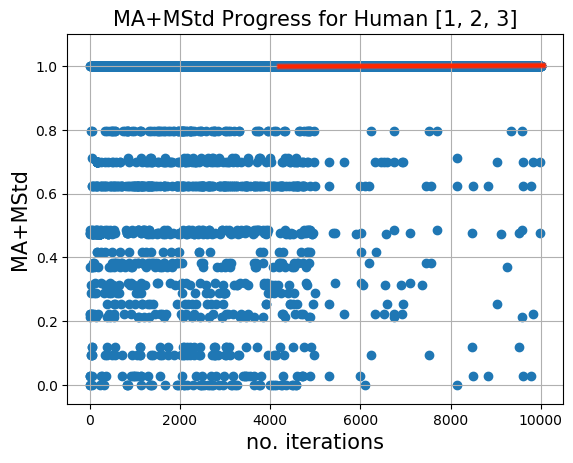}}
    \caption{}
    \label{fig:nocvrewardprogressthermal}
    \end{subfigure}
    &
    \begin{subfigure}{0.22\textwidth} 
    \raisebox{-0.3\totalheight}{\includegraphics[width=\textwidth]{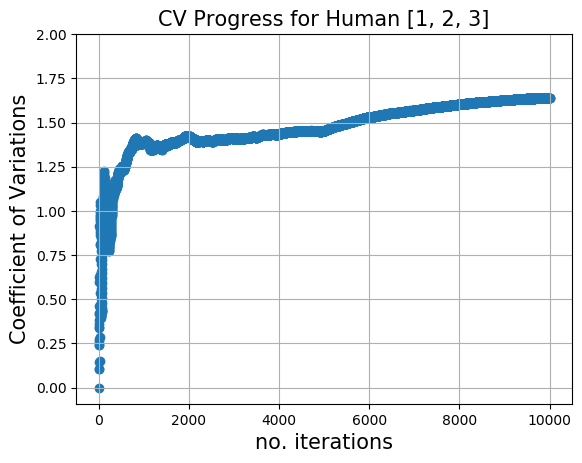}}
    \caption{}
    \label{fig:nocvrewardprogressthermal} 
    \end{subfigure}
     \\ 
     \rotatebox{90}{\textbf{With fairness}}
     &
     \begin{subfigure}{0.2\textwidth} 
     \raisebox{-0.5\totalheight}{\includegraphics[width=\textwidth]{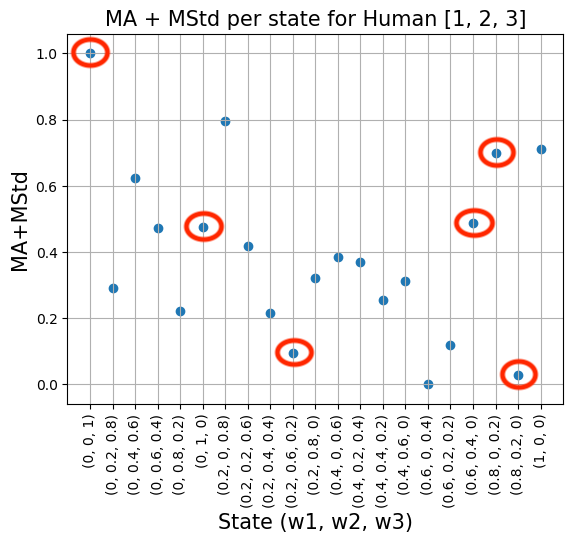}}
     \caption{}
     \label{fig:withcvmamstd} \end{subfigure}
     &
    \begin{subfigure}{0.25\textwidth} 
    \raisebox{-0.5\totalheight}{\includegraphics[width=\textwidth]{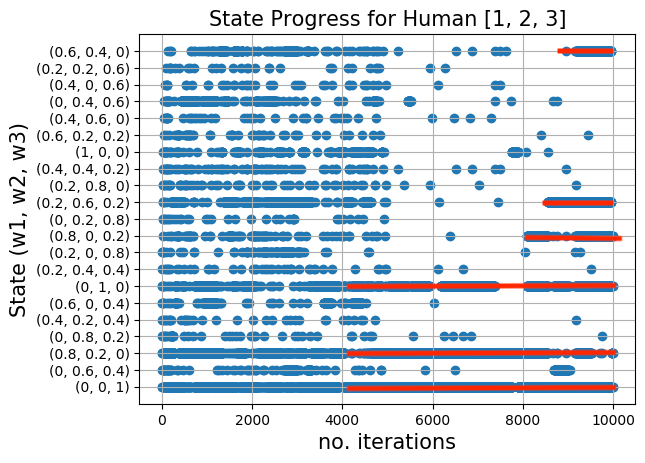}}
     \caption{}
     \label{fig:withcvstateprogressthermal} \end{subfigure}
     &
    \begin{subfigure}{0.22\textwidth} 
    \raisebox{-0.5\totalheight}{\includegraphics[width=\textwidth]{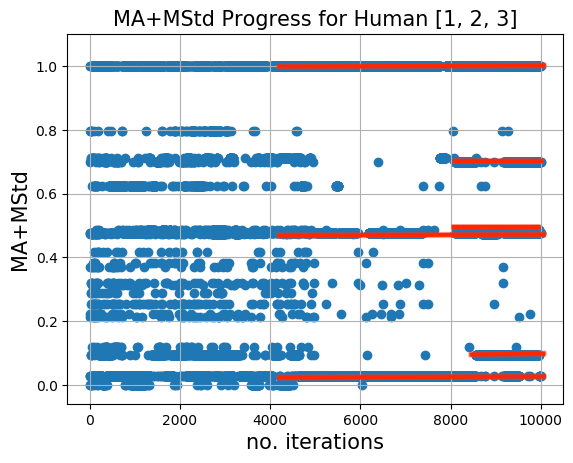}}
     \caption{} 
     \label{fig:withcvrewardprogressthermal} \end{subfigure}
    &
    \begin{subfigure}{0.22\textwidth} 
    \raisebox{-0.5\totalheight}{\includegraphics[width=\textwidth]{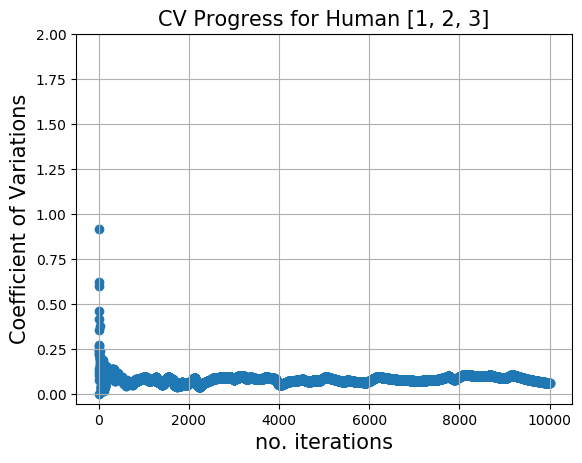}}
    \caption{}
    \label{fig:withcvcvrewardprogressthermal} \end{subfigure}
\end{tabularx}
\caption{Numerical results of Experiment 4 and 5 by comparing the performance of the Mediator RL without/with the fairness measure. (a) and (e) show the performance of the system---obtained through brute force search---under different assignments for $(w_1,w_2,w_3)$. The states that the Mediator RL chooses are circled. (b) shows the progress of the states of the Mediator RL agent without the fairness measure $\mathcal{F}$ across different iterations of the system showing the convergence to the state with maximum system performance, while (f) shows the states that Mediator RL chooses with fairness measure which are multiple states that correspond to the circled states in (e). While (c) shows the performance of the system due to the adaptations of both the Mediator RL and the Governor RL agents showing an increase in the performance over time, while (g) shows that with fairness measure, the performance of the system depends on the states that the Mediator RL chooses to balance the fairness across the humans. (d) and (h) show the difference in the coefficient of variation. The low values indicate higher fairness.}
\label{fig:exp4}
\vspace{-2mm}
\end{figure*}

\begin{figure*}[!th]
\centering
    \begin{tabularx}{\textwidth} {  p{0.22\textwidth}  p{0.22\textwidth} | p{0.22\textwidth}  p{0.22\textwidth}} 
    \multicolumn{2}{c|}{\textbf{Fixed set-point}} & 
    \multicolumn{2}{c}{\textbf{FaiR-IoT}} \\\hline 
     \includegraphics[width=0.25\textwidth, scale=0.5]{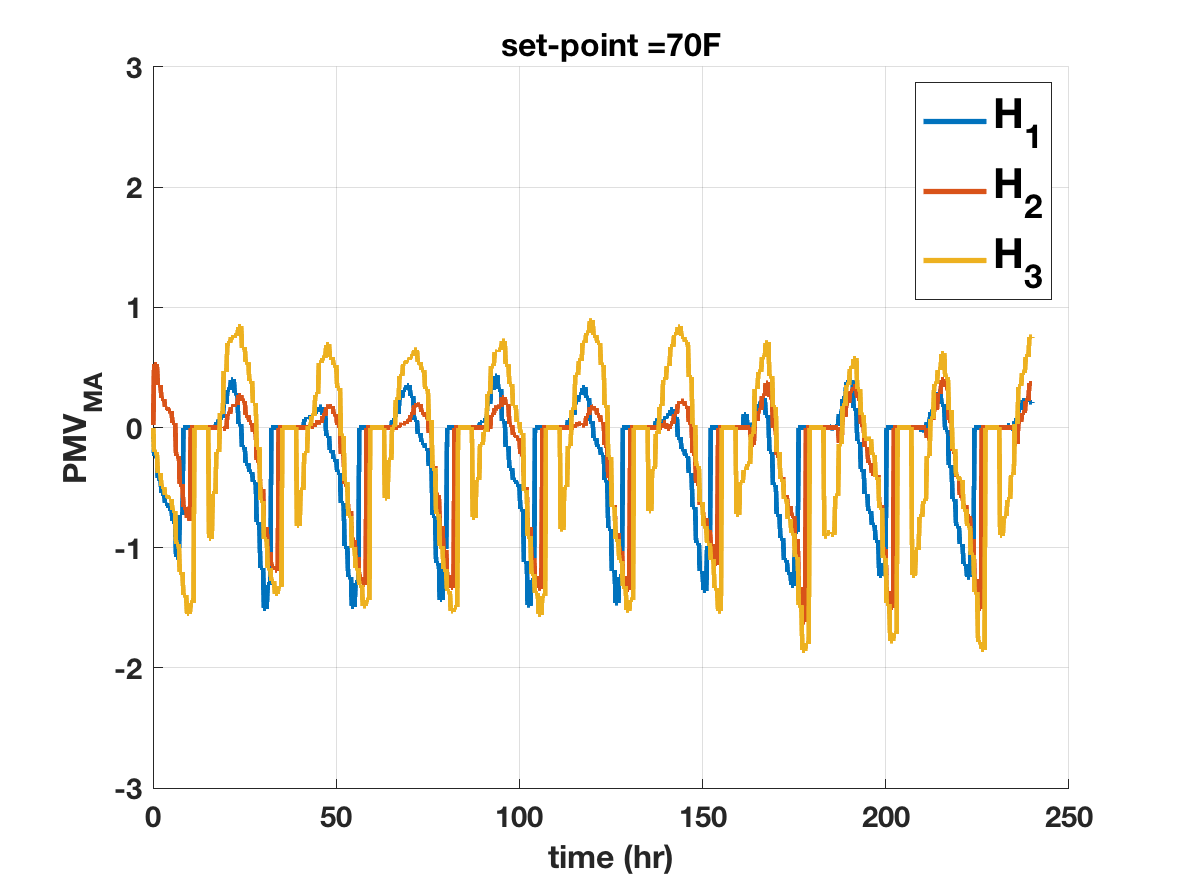} &
     \includegraphics[width=0.25\textwidth, scale=0.5]{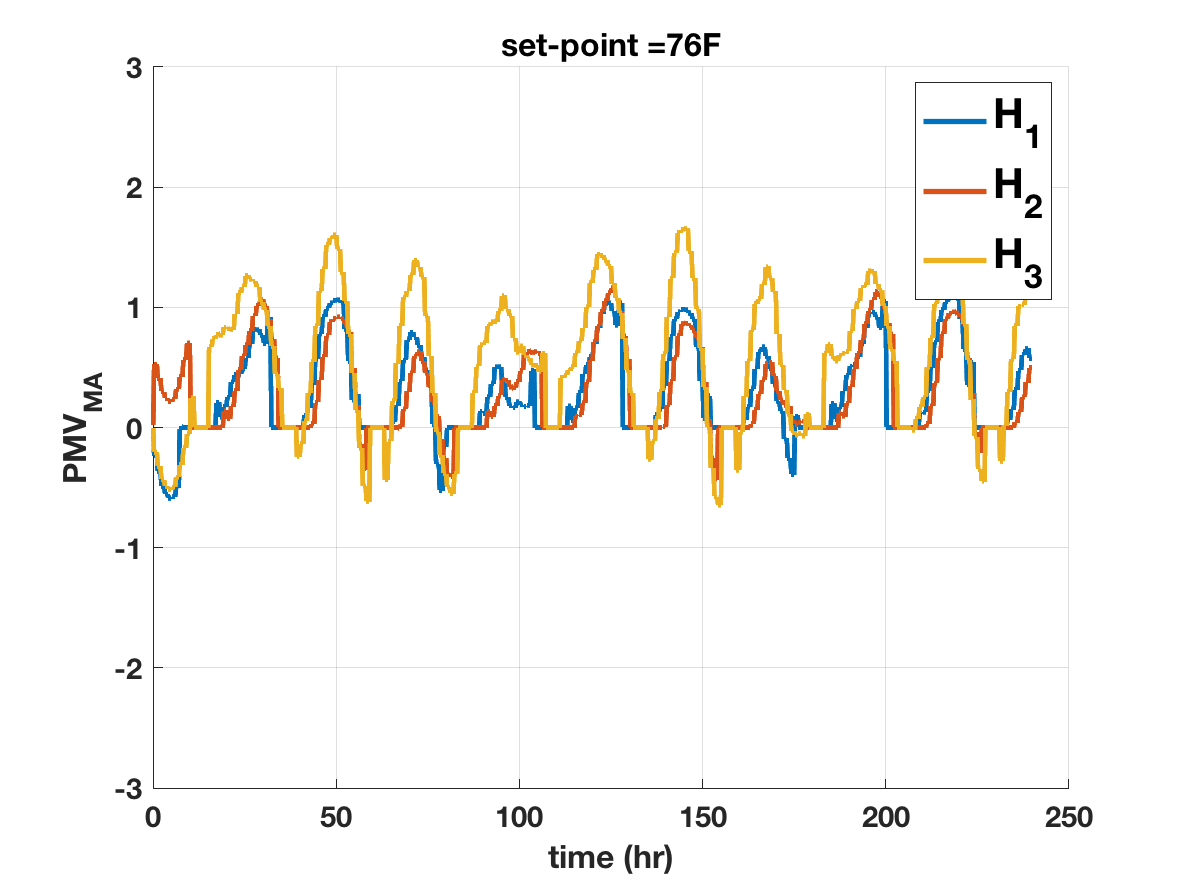} &
     \includegraphics[width=0.25\textwidth, scale=0.5]{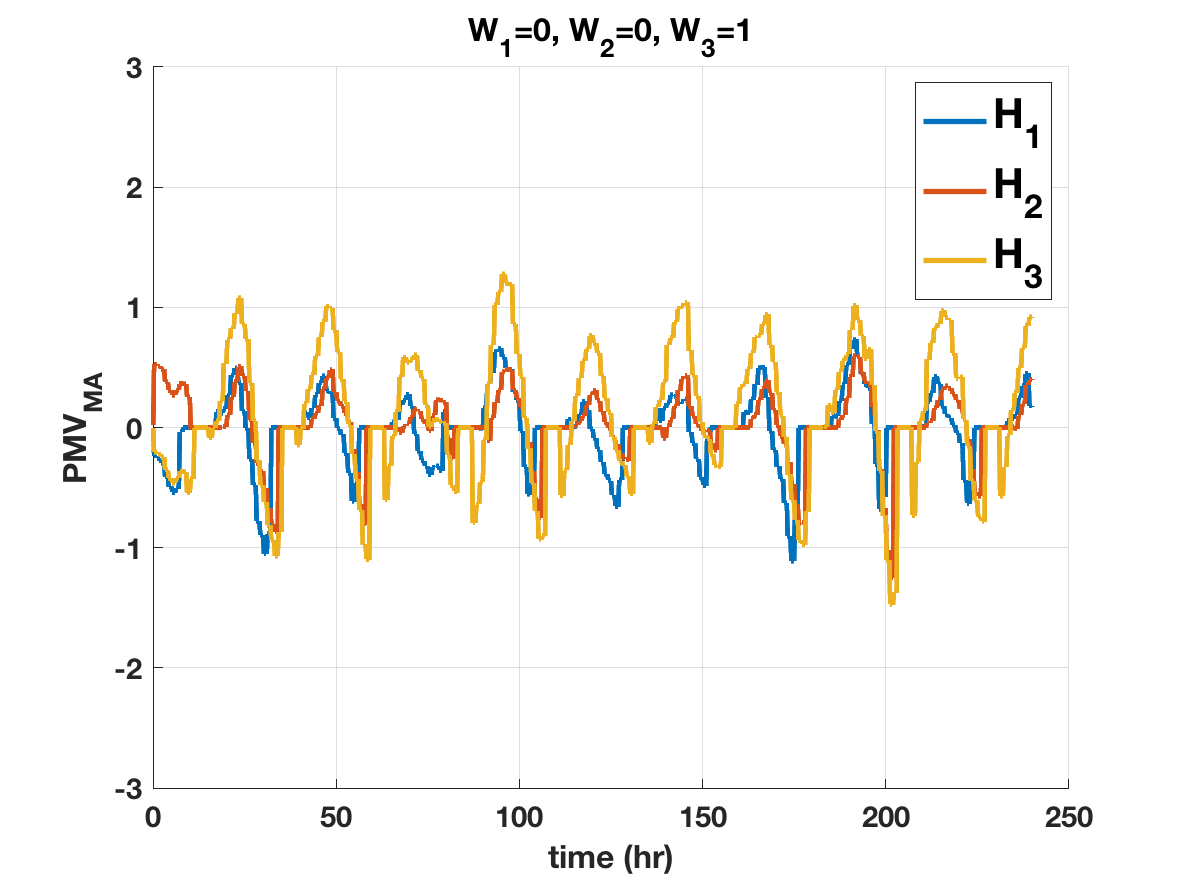} &
     \includegraphics[width=0.25\textwidth, scale=0.5]{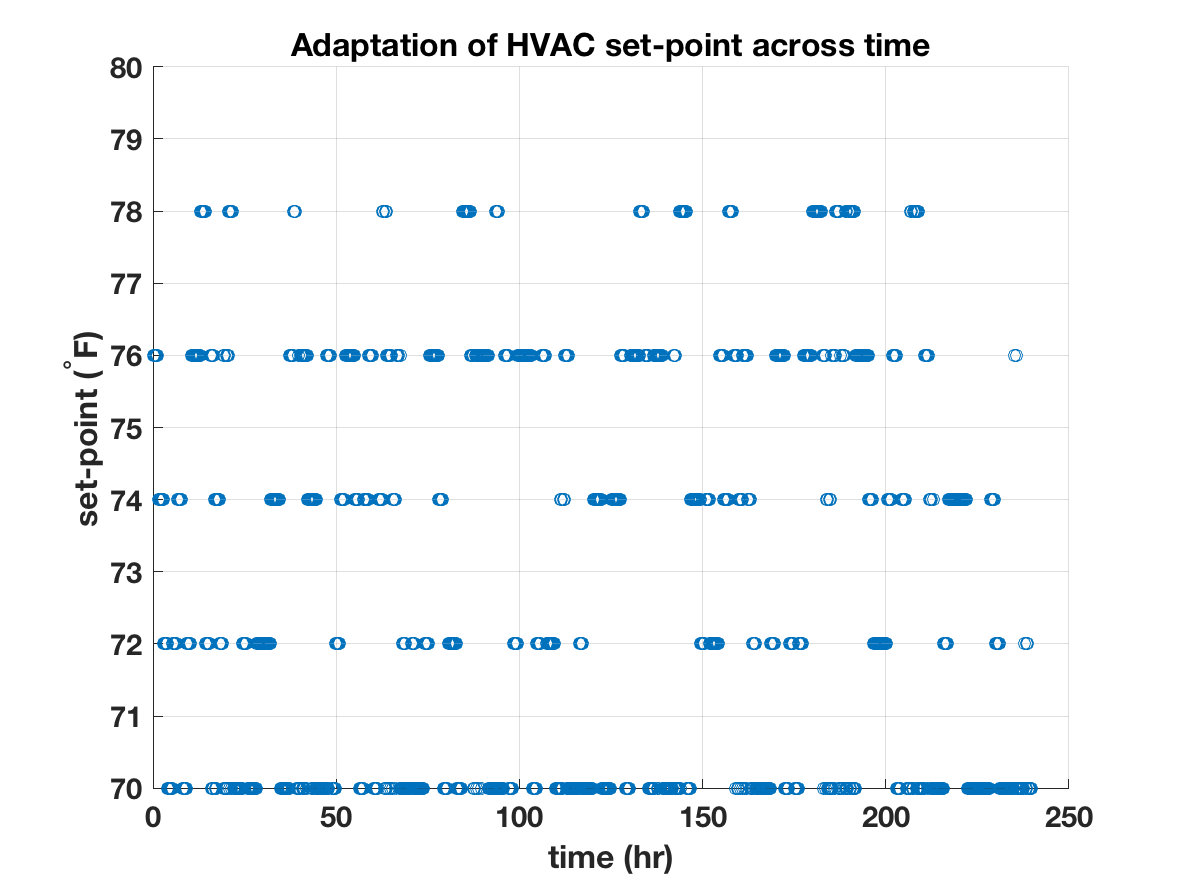}
     \end{tabularx}
\caption{Numerical results of Experiment 6 comparing the PMV of the three residents when: Moving Average of PMV for the three humans at fixed HVAC set-point of $70F$ and $76F$ and Moving average of PMV for the three humans when the proposed \sysname framework is used to adapt the HVAC set-point along with the computed set points. Thanks to the proposed \sysname framework, the PMV of the three residents is maintained within the acceptable thermal comfort of [-1,1].}
\label{fig:exp5}
\vspace{-3.5mm}
\end{figure*}

\subsection{Experiment 4: Multi-Human Variability}\label{sec:exp4}
Now, we study the ability of the proposed system (with both Governor RL and Mediator RL agents interacting together) to provide personalized performance when multiple occupants are present in the system. Similar to the previous experiments, we start by brute-forcing all the states of the Governor and the Mediator RL agents to obtain the ground truth for the performance of the system under different values for $T_l, T_a$ for each human along with the different values for the mediator weights $w_1, w_2, w_3$. While the optimal values for $T_l, T_a$ for each human (individually) was discussed in Experiment 3, we report in Figure~\ref{fig:nocvmamstd} the system performance for different values of $w_1, w_2, w_3$. In particular, state (0, 0, 1) shows the best performance while states (0.6, 0, 0.4) and (0.8, 0.2, 0) show the worst performances. 
Next, we run the whole system and compares its convergence to the state with the maximum total reward. First we run the whole system \emph{without} fairness in Mediator RL reward by assigning $\zeta$ to value $zero$ as discussed in Section~\ref{sec:fair-mediator}. 

The progress of the states is shown in Figure~\ref{fig:nocvstateprogressthermal}, where the Mediator RL converges to (0, 0, 1) at iteration 3000, which is the state of maximum performance. This is also reflected in how the performance value changes in Figure~\ref{fig:nocvrewardprogressthermal}. 

This is a counter-intuitive result since it entails that the system's optimal performance is achieved when the HVAC set-point ultimately favors human $H_3$. However, one explanation for this result is that human $H_3$ stays in the house for more time than both $H_1$ and $H_2$ and hence the moving average reward is maximized when human $H_3$ is given the highest priority. This motivates using the fairness-aware Mediator RL as explained in Section~\ref{sec:fair-mediator} for the next experiment.
\vspace{-1mm}
\subsection{Experiment 5: Fairness with Multi-Human Variability}\vspace{-1mm}
We run the same experiment discussed in experiment 4 but with the fairness-aware Mediator RL by assigning $\zeta$ a value of $0.5$ to balance the reward between the performance measure $\mathcal{W}$ and the fairness measure $\mathcal{F}$. As seen in Figure~\ref{fig:withcvstateprogressthermal}, the states that the Mediator RL chooses were a combination of different states across time. In particular, states (0, 0, 1), (0, 1, 0), (0.2, 0.6, 0.2), (0.6, 0.4, 0), (0.8, 0, 0.2), and (0.8, 0.2, 0) were selected across time. This shows that the Mediator RL changes the priority of adaptation between $H_1$, $H_2$, and $H_3$ across time to ensure fairness. As a result, the performance does not converge to the maximum value state, but it changes depending on the current weights assigned by the Mediator RL. However, even though the performance does not converge to the maximum value, the fairness increases. This is reflected by the difference in the coefficient of variation ($cv$) values between experiment 4 (Figure~\ref{fig:nocvrewardprogressthermal}) and experiment 5 (Figure~\ref{fig:withcvcvrewardprogressthermal}). The lower value of $cv$ indicates higher fairness. In particular, the $cv$ has been improved by $1.5$ orders of magnitude through integrating the fairness measure $\mathcal{F}$ in the reward function of the Mediator RL.

\vspace{-1mm}
\subsection{Experiment 6: Personalized vs Static Smart HVAC Systems}\vspace{-1mm}
Finally, we assess the value of personalizing the experience of the smart HVAC system compared to non-personalized HVAC that uses fixed, manually chosen set-point. Figure~\ref{fig:exp5} (left) shows the PMV of all three simulated humans for fixed set-points of $70F$ and $76F$, respectively. These figures show that the PMV of the three residents exceeds the acceptable range of comfortable thermal, which is between -1 and 1~\cite{handbook2009american}. On the other hand, and as shown in Figure~\ref{fig:exp5}(right), the PMV of the three residents exhibits a more favorable behavior due to the set-point adaptations provided by the proposed framework. It is worth noting that the personalized framework does utilize these two set-points $70F$ and $76F$ most of the time (as shown in Figure~\ref{fig:exp5} (right)). Nevertheless, thanks to its ability to switch between these two values depending on the resident comfort, the system's performance is improved by $41.7\%$ and $58.96\%$ compared to the manually fixed set-point scenarios.

\section{Conclusion}
Human modeling and human preference prediction hold the promise of disrupting the status-quo by designing complex IoT systems that weave advances in human sensing into the fabric of large-scale and societal IoT systems. However, with multiple humans interacting within the same IoT application space, adaptation fairness is a crucial concern. In this paper, we proposed \textbf{\sysname}, a Fairness-aware Human-in-the-Loop Reinforcement Learning framework that addresses the challenge of human variability modeling with the fairness of adaptation. \sysname uses three hierarchical RL agents; a Governor RL with a Multisample RL to address the intra-human and inter-human variability, and a fairness-aware Mediator RL to address the multi-human variability. We showed the ability of \sysname to personalize two different IoT applications in the domain of ADAS and smart homes. By adapting to the human's variability, \sysname was able to improve the human experience by 40\% to 60\% compared to the non-personalized systems and enhancing the system's fairness by 1.5 orders of magnitude. Given that this framework focuses on establishing the intersection between human modeling and fairness in IoT applications, it opens up new research and application development directions for more fair and human-centric IoT.


\bibliographystyle{ACM-Reference-Format}
\bibliography{ref.bib}

\end{document}